\documentclass{article}

    \PassOptionsToPackage{numbers, compress}{natbib}
\usepackage[preprint]{neurips_2022}

\usepackage[utf8]{inputenc} % allow utf-8 input
\usepackage[T1]{fontenc}    % use 8-bit T1 fonts
\usepackage{hyperref}       % hyperlinks
\usepackage{url}            % simple URL typesetting
\usepackage{booktabs}       % professional-quality tables
\usepackage{amsfonts}       % blackboard math symbols
\usepackage{nicefrac}       % compact symbols for 1/2, etc.
\usepackage{microtype}      % microtypography
\usepackage{xcolor}         % colors
\usepackage{natbib}
\usepackage{amsmath}
\usepackage{graphicx}
\usepackage{subfig}
\usepackage{multirow}
\usepackage{float}
\usepackage{comment}
\usepackage{wrapfig, blindtext}
\usepackage{ulem}
\title{Efficient LiDAR Point Cloud Geometry Compression Through Neighborhood Point Attention}

\author{%
	Ruixiang Xue, Jianqiang Wang, Zhan Ma\\
	Nanjing University\\
	\texttt{ \{xrxee, wangjq\}@smail.nju.edu.cn}, \texttt{mazhan@nju.edu.cn} \\
}

\begin{document}

\maketitle

\begin{abstract}
Although convolutional representation of  multiscale sparse tensor demonstrated its superior efficiency to accurately model the occupancy probability for the compression of geometry component of dense object point clouds, its capacity for representing sparse LiDAR point cloud geometry (PCG) was largely limited. This is because 1) fixed receptive field of the convolution cannot  characterize extremely and unevenly distributed sparse LiDAR points very well; and 2) pretrained convolutions with fixed weights are insufficient to dynamically capture information conditioned on the  input. This work therefore suggests the {\it  neighborhood point attention} (NPA) to tackle them, where we first use $k$ nearest neighbors ($k$NN) to construct adaptive local neighborhood; and then leverage the self-attention mechanism to dynamically aggregate information within this neighborhood. Such NPA is devised as a NPAFormer to best exploit cross-scale and same-scale correlations for geometric occupancy probability estimation.
%apply the $k$NN ($k$ nearest neighbors) clustering to agg 
%We specifically answer: how to define and use the neighborhood information dynamically.
%One is that  are hard to be characterized using stacked convolutions with fixed
%The work is an improvement of the previous proposal SparsePCGC presented in [1]. SparsePCGC is a point cloud geometry compression framework using sparse tensor-based multiscale representation, which demonstrates strengths in lossless and lossy compression of different point clouds.
%In this document, we further optimize the implementation of the original version and introduce the local attention mechanism based on K Nearest Neighbors (KNN) to aggregate neighboring points, which is more efficient than the pure convolution network in sparse LiDAR data compression. The designed attention layer can be easily embedded in the multiscale and multistage structure of SparsePCGC. 
Compared with the anchor using standardized {\it G-PCC}, our method provides  {$>$17\%} BD-rate gains for lossy compression,  and $>$14\% bitrate reduction for lossless scenario using popular LiDAR point clouds in SemanticKITTI and Ford datasets. Compared with the state-of-the-art (SOTA) solution using attention optimized octree coding method,  our approach requires much less decoding runtime with about {640$\times$} speedup on average, while still presenting better compression efficiency.
\end{abstract}

\section{Introduction} \label{sec:intro}

LiDAR sensor becomes one of many commodity devices equipped on vehicles, robots and drones for autonomous machinery through the enabling of high-precision geometry perception using dynamically-generated 3D LiDAR point clouds~\cite{Schwarz2019EmergingMS}.  However, efficient compression of massive LiDAR points that is vital for the storage and networked exchange in vast applications~\cite{Chen_PC_automotive} is still challenging because it is hard to exploit inter-correlations among unstructured sparse points. To tackle it, numerous 3D representation models like uniform voxel~\cite{Wang2021Lossy,Guarda2021AdaptiveDL,Quach2020ImprovedDP}, octree~\cite{meagher1982geometric}, multiscale sparse tensor~\cite{Wang2020MultiscalePC,wang2021sparse}, etc, are developed  to specify  explicit  neighborhood connections upon which rules or learning based approaches~\cite{MPEG_PCC_PIEEE,quach2022survey}  exploit inter-dependency among points in close proximity. 

Among them, convolutional representation of multiscale sparse tensor, noted as {\it SparsePCGC}~\cite{wang2021sparse}, has demonstrated encouraging performance on the compression of geometry component for both dense object and sparse LiDAR point clouds.  This is mainly because of the utilization of multiscale and multistage processing to effectively exploit cross-scale and same-scale correlations for accurate geometric occupancy probability approximation. To this aim, {\it SparsePCGC} stacks  3D sparse convolutions (SConv) for neighborhood information characterization and embedding. However, the compression gain of LiDAR point cloud geometry (PCG)  is relatively marginal when compared with that of dense object PCG. We believe this is because, the neighborhood correlation among  extremely and unevenly distributed sparse LiDAR points is hard to model by stacked SConvs that come  with the fixed receptive field and fixed weights (after training).

%incapable of aggregating effective information.    

%that are dynamically and sparsely distributed in 3D space 

%As an important branch of AI-PCC, sparse LiDAR data has been deeply studied in recent research. Currently, most of the related work is based on the learned octree entropy coding, such as OctSqueeze [2], VoxelContextNet [3], PointContextNet [4], OctAttention [5], etc. Besides, in our previous proposal [1], we presented a sparse tensor-based PCGC framework SparsePCGC, which can be applied to both dense objects and sparse LiDAR point clouds, showing a wide range of application scenarios and competitive compression performance.

%However, SparsePCGC mainly uses stacked sparse convolution layers to aggregate neighboring points, which works well for dense point clouds, but may not be efficient for sparser point clouds. 

\begin{figure*}[t]
  \subfloat[]{\includegraphics[scale=0.20]{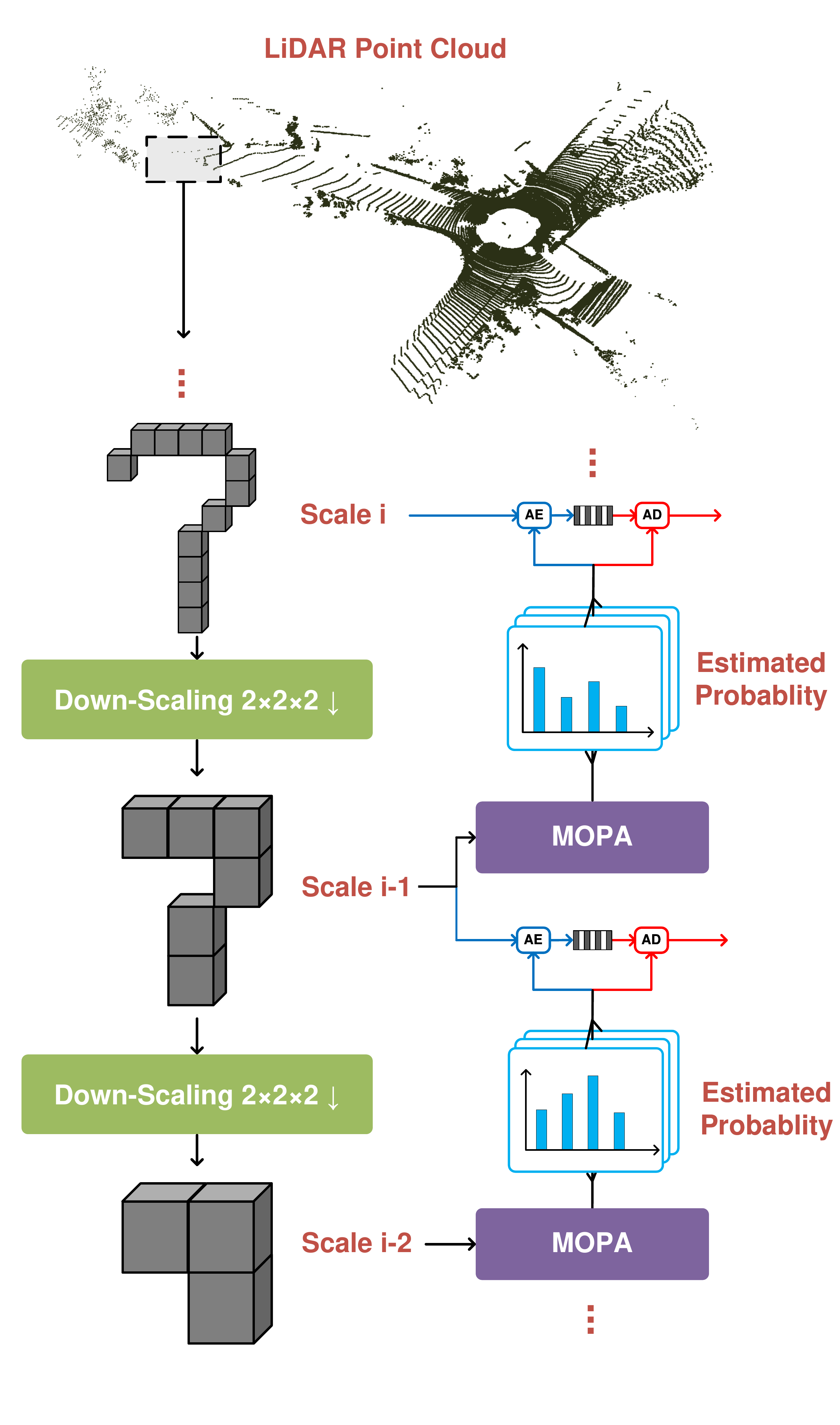} \label{sfig:multistage_pipeline}}
  \subfloat[]{\includegraphics[scale=0.20]{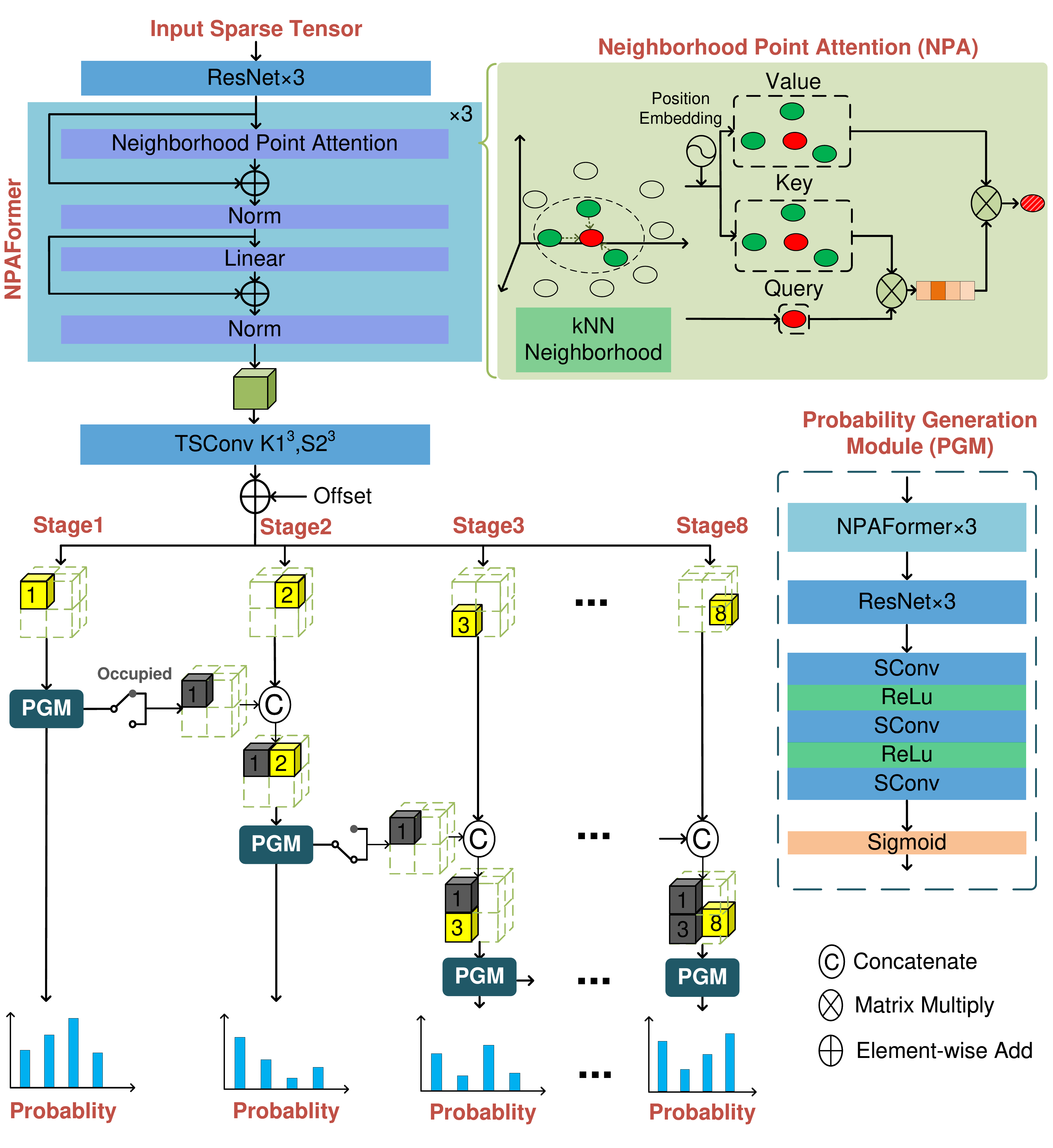}\label{sfig:mopa}}
\caption{{\bf Framework.} (a) Multiscale Sparse Tensor Representation exploits  cross-scale and same-scale correlations exhaustively using MOPA (Multistage Occupancy Probability Approximation). Estimated probability is used to perform the entropy coding and decoding of geometric occupancy. (b) Each MOPA inputs the sparse tensor of positively occupied voxels (POVs) from preceding scale, e.g., ($i-1$)-th, to produce occupancy probabilities for encoding/decoding at current  scale, e.g., $i$-th, where NPAFormer first aggregates neighborhood information at ($i-1$)-th scale for each POV,  then TSConv upscales each POV (w/ aggregated neighborhood features) to eight child nodes (octant) in parallel and then PGM (Probability Generation Module) stagewisely outputs occupancy probability and determines the occupancy status of  each octant. AE and AD are for arithmetic encoding and decoding. SConv is for sparse convolution, and TSConv is transposed SConv. }
\label{fig:framework}
\end{figure*}

We therefore suggest the {\it Neighborhood Point Attention} (NPA) to tackle aforementioned issues. As in Fig.~\ref{sfig:multistage_pipeline}, we follow~\cite{wang2021sparse} to use  multiscale sparse tensors to process the collection of positively occupied voxels (POVs) from one scale to another, where dyadic downscaling  at all axes (e.g., $x$-, $y$- and $z$-axis) in Cartesian coordinate  system is used to generate multiscale representations. Such dyadic downscaling is the same as the  depth adaptation used in~\cite{Huang_2020_CVPR,Que_2021_CVPR,fu2022octattention} to squeeze and build parent-child octree structure for correlation exploration. %, making the comparisons much easier with them by just using the {\it bits per point} (bpp). 

Different from the {\it SparsePCGC}~\cite{wang2021sparse} that thoroughly relies on stacked SConvs to exploit correlations for occupancy probability estimation. This work however devises a novel MOPA (Multistage Occupancy Probability Approximation) to fulfill this purpose, where the MOPA accepts the sparse tensor of POVs from preceding scale (e.g., $(i-1)$-th), and outputs the probability of each element of upscaled sparse tenor at current scale (e.g., $i$-th) for encoding or decoding as in Fig.~\ref{sfig:multistage_pipeline}.  %Note that the MOPA process traverses all POVs in input sparse tensor.

As illustrated in Fig.~\ref{sfig:mopa}, for each POV, we first build a local neighborhood using its $k$ nearest neighbors ($k$NN), and then execute the self-attention computation in this  neighborhood. Such NPA is implemented as a NPAFormer for local information aggregation and embedding. {The POVs} with aggregated neighborhood features are then upscaled and added a set of offsets to generate eight child nodes (octants) in parallel. After then, the PGM (Probability Generation Module) that stacks NPAFormer and SConv layers stagewisely processes the octant to produce occupancy probability for compression at current scale. %Note that the MOPA allows parallel processing at each step thoreticaly, but our current implementation does not leverage this 

%and plugged into the multiscale sparse tensor representation framework to best exploit cross-scale and same-scare correlation for accurate occupancy probability  approximation (OPA) of LiDAR PCG compression.

%best characterize the dynamics and sparsity of LiDAR PCG for better compression. To this aim, 
As seen, the use of $k$NN neighborhood in NPAFormer relaxes the receptive field restriction of stacked SConvs used in {\it SparsePCGC}~\cite{wang2021sparse} by flexibly including more valid and effective POVs nearby; while in contrast, fixed convolutional kernel presets like $3\times3\times3$ sometimes may only contain very few valid POVs (e.g., 1 or 2) especially in regions having relatively-low point density~\cite{zhu2021lossy}, which largely lacks sufficient information for accurate probability estimation. Subsequently $k$ neighbors are aggregated using self-attention to weigh the contribution dynamically conditioned on the input neighborhood. Note that $k$NN neighborhood varies accordingly for each specific point, well reflecting the content dynamics, which  makes the information embedding more robust and efficient~\cite{lu2021transformer}.

%Thus in this document, we design a novel transformer layer based on the local attention mechanism to aggregate the information of K nearest neighboring points, which is inspired by the success of transformers and self-attention networks in point cloud processing and data compression. The designed transformer network can serve as the backbone like CNNs to build the multiscale and multistage structure like in SparsePCGC.

Extensive experiments have reported the superior  efficiency of the proposed method for compressing LiDAR PCGs like SemanticKITTI and Ford sequences in various settings. Having the  anchor using standard compliant {\it G-PCC} (Geometry based Point Cloud Compression)~\cite{MPEG_GPCC_CTC,MPEG_PCC_PIEEE}, multiscale sparse tensor representation using the proposed NPA shows  $>$17\% BD-rate (Bj\o ntegaard Delta rate) gains~\cite{BDrate} for lossy compression, and $>$ 14\% compression ratio improvement in lossless modes, which can be translated to about 10 absolute percentage points improvement over the respective gains obtained by the {\it SparsePCGC}~\cite{wang2021sparse} that applies SConvs upon the same multiscale sparse tensor. In the meantime, compared with the SOTA solution using attention optimized octree coding model, a.k.a., {\it OctAttention}~\cite{fu2022octattention}, our method not only provides compression improvement, particularly at high bitrates that is more preferred by high-precision tasks in autonomous driving (e.g., {$\approx$7\%} BD-rate gains from 6 to 16 bpp), but also enormously reduces the decoding complexity to several orders of magnitude, e.g., less than 6 seconds of the proposed method  vs. $\approx$1 hour  of the {\it OctAttention} when decoding SemanticKITTI sequences (about 640$\times$ speedup on average in Table~\ref{tab:encoding_decoding_time}).

%the designed neighborhood attention transformer significantly outperforms the original convolution layers in the compression of LiDAR point clouds, which brings > 10\% BD-Rate gains in various test conditions, with even less runtime. In future research, we will further explore its potential value in other tasks.
%Please read the instructions below carefully and follow them faithfully.

{\bf Contributions} of this work include: 1) We show that neighborhood point attention that dynamically characterizes and embeds information among $k$ nearest neighbors can better aggregate effective information than the fixed-weights and fixed-receptive-field convolutions; 2) Together the multiscale sparse tensor representation, the proposed NPA mechanism can better exploit cross-scale and same-scale correlations with the SOTA efficiency on the compression of sparse LiDAR PCGs; 3) Having the computation of NPA on $k$ nearest neighbors significantly reduces the complexity compared with global self-attention, promising attractive prospects for practical applications.

%to effectively compress the geometry of sparse LiDAR point clouds, for which $k$ nei; 

\section{Related Work} \label{sec:related_work}

\textbf{Explicit 3D Representation Models of LiDAR PCG.} Uniform voxel model is the most straightforward way for PCG representation, but it is more suitable for dense object PCG~\cite{Wang2021Lossy,Guarda2021AdaptiveDL,Quach2020ImprovedDP}. Octree model is a lightweight and effective approach through the use of parent-child tree decomposition, with which the compression efficiency is improved by carefully exploiting parent-child dependency. Notable coding solutions using octree include standardized {\it G-PCC} that applies  rules based contexts, and  {\it OctSqueeze}~\cite{Huang_2020_CVPR}, {\it MuSCLE}~\cite{biswas2020muscle}, {\it VoxelContextNet}~\cite{Que_2021_CVPR}, {\it OctAttention}~\cite{fu2022octattention}, etc that utilizes neural network based solutuions like CNN (Convolutional Neural Network), MLP (Multi-Layer Perceptron), and Transformers.  Recently, a promising multiscale sparse tensor representation using stacked SConvs has emerged~\cite{Wang2020MultiscalePC,wang2021sparse} with great flexibility to exploit cross-scale and same-scale correlations. Although the compression gain of multiscale sparse tensor on LiDAR PCGs is relatively marginal~\cite{wang2021sparse}, we believe it is mainly because of the inefficient characterization and embedding of neighborhood points for redundancy removal. %This  

\textbf{Sparse Tensor.} LiDAR point cloud is comprised of a collection of sparsely and unevenly distributed points which inherently can be represented using a sparse tensor consisting of  a set of POV coordinates $C = \{(x_i,y_i,z_i)\}_i\in\mathbb{R}^{N \times 3}$ and associated features $F=\{f_i\}_i\in\mathbb{R}^{N \times d}$.  In practice, sparse tensor can be easily implemented using Minkowski Engine~\cite{choy20194d}, where a hash map is applied to index POV coordinates and an auxiliary data structure is also included to manage their connections efficiently. Aforementioned multiscale sparse tensor can be fulfilled by spatially resampling~\cite{Wang2020MultiscalePC}. 
Note that  only POVs are involved in computations like convolution, leading to attractive lightweight solution. 

\textbf{Sparse Convolution.} Upon the sparse tensor, it is naturally to apply the SConv as specified in~\cite{choy20194d}:
\begin{equation}
\vspace{-0.025in}
f_{u}^{out} = \sum\nolimits_{i \in \mathbb{N}^{3}(u, C^{in})} W_{i} f_{u+i}^{in} 
\quad\text{for}\quad
u \in C^{out}, \label{eq:sparse_conv}
\end{equation} 
%\vspace{-0.025in}
having $C^{in}$ and $C^{out}$ as the coordinate sets for input and output POVs.  
%which are not necessarily the same, leading to the support of flexible re-sampling.
$f_{u}^{in}$ and $f_{u}^{out}$ are input and output feature vectors at coordinate $u$. 
$\mathbb{N}^{3}(u, C^{in}) = \{i|u+i \in C^{in}, i\in \mathbb{N}^{3}\}$ defines a 3D convolutional kernel with a predefined size like $3\times3\times3$ or $5\times5\times5$, covering a set of locations centered at $u$ with offset $i$ in $C^{in}$. $W_i$ denotes the corresponding kernel weight at offset $i$ centered at $u$. As seen, the neighborhood coverage for information aggregation is basically  constrained by the kernel size of convolutions; and as in \eqref{eq:sparse_conv}, convolutional weights $W_i$s are fixed after training, which clearly is incapable of effectively characterizing the dynamics unseen in training.

\textbf{Self-attention \& Transformer.} 
%Inspired by the success of self-attention layer and transformer architectures\cite{vaswani2017attention} in the NLP field,some works\cite{dosovitskiy2020image,liu2021swin} employ self-attention layers to replace some or all of the 2D convolution layers and achieve better performance than convolutional neurl networks.In the implementation of Attention function,the input are query and key-value pairs,and the ouput is a wighted sum of values,where the attention weights are computed by query and corresponding key.Its formula can be expressed as follows:
%\begin{equation}
 %   Attention(Q,K,V)=softmax(\frac{QK^{T}}{\sqrt{d_{k}}} )V 
%\end{equation}
%where $d_{k}$ is a scaling factor that equals to the key dimension.
Recently, self-attention mechanism and Transformers are migrated quickly to process point clouds, showing encouraging results for different task including segmentation, classification, compression, etc~\cite{Guo_2021,zhao2021point,engel2021point,fu2022octattention}. However, existing solutions often demand huge computational cost and memory consumption that grow quadratically with the sequence length of underlying tokens.  This complexity issue is of particular importance for practical applications. This work thus examines the use of neighborhood point attention to pursue the advantages from both local processing (e.g., lightweight complexity) and self-attention weighting (e.g., dynamic information embedding).

%transformer architecture as an efficient neighborhood embedding method.This architecture localizes the receptive field to K nearest neighbors around each query.We combine this architecture with sparse tensor based multiscale representation point cloud geometry compression framework\cite{wang2021sparse} and achieve competitive result in LiDAR point cloud geometry compression. 

\section{Method}
\subsection{Multiscale Sparse Tensor with MOPA}
Figure~\ref{sfig:multistage_pipeline} illustrates the LiDAR PCG compression framework using multiscale sparse tensor with the proposed MOPA to accurately estimate the occupancy probability for encoding and decoding. As aforementioned, dyadic downscaling is enforced to derive multiscale sparse tensors where scale-wise sparse tensor can be easily mapped to depth-wise octree-structured representation as well. 
Different from those octree coding approaches~\cite{Huang_2020_CVPR,fu2022octattention} that exploit correlations following the parent-child tree structure, we leverage cross-scale and same-scale correlations through the use of MOPA for compression. %To ensure the general applicability of the MOPA, we best train a single MOPA model for the use in a given  bitrate range.

\subsubsection{MOPA - Multistage Occupancy Probability Approximation}

Figure~\ref{sfig:mopa} details the MOPA used in this work. In MOPA, it combines the NPA and SConv layers (but not just SConvs as in {\it SparsePCGC}~\cite{wang2021sparse}). Note that NPA Transformer is devised using the popular Transformer architecture for processing, dubbed as the NPAFormer for simplicity.
%to input the sparse tensor of preceding scale e.g., $(i-1)$-th, to generate occupancy probability of current scale, e.g., $i$-th for encoding and decoding. 

Assuming the input sparse tensor at $(i-1)$-th scale as $\mathbb{S}_{i-1}$, the propose MOPA processes all POVs in it to generate the occupancy probability of each element  of upscaled sparse tensor $\mathbb{
\tilde{S}}_{i}$ at $i$-th scale. Note that $\mathbb{S}_{i-1}\subset\mathbb{
\tilde{S}}_{i}$ because the elements in $\mathbb{
\tilde{S}}_{i}$ include both POVs and non-POVs. Whether this element is a POV or non-POV is fully  known in encoder for encoding, while the status of POV or non-POV in decoder is determined by parsing the bitstream, through the use of occupancy probability respectively.

The MOPA includes three major steps. As for the processing from $(i-1)$-th  to $i$-th scale, 
\begin{itemize}
\item First, it stacks the ResNet~\cite{he2016deep} and NPAFormer to aggregate neighborhood information at $(i-1)$-th scale where $3\times3\times3$ SConvs are used in ResNet, and NPAFormer applies the NPA upon $k$ nearest neighbors chosen by the Euclidean distance on-the-fly; Note that all POVs in  $\mathbb{S}_{i-1}$ can be processed concurrently.% in theory.
\item Then, each POV with aggregated neighborhood features in $\mathbb{S}_{i-1}$ is upscaled using transposed SConv, e.g., ``TSConv K1$^3$, S2$^3$'', with  convolutional kernel at a size of 1$\times$1$\times$1 and upscaling stride  of 2 at three axis dimensions, and added corresponding offset to directly generate eight child nodes (yellow cubic with number mark); Such operation can be executed in parallel for all POVs from $(i-1)$ scale. 
\item Third, a PGM that combines the NPAFormer and SConv layers is devised to stagewisely estimate the occupancy probability for each child node, where the probability approximation of succeeding child node (e.g., node ``3'' or ``2'' in yellow cubic) also includes the occupancy status of proceeding child node (e.g., node ``1'' as POV in dark grey cubic).  Note that ReLU (Rectified linear unit) is coupled with SConv for non-linearity, and Sigmoid layer is used for probability derivation in the range of [0,1]. And it is also worth to point out that stage-wise PGM is made concurrently for same-group child nodes  at $i$-th scale that are labelled with the same number but upscaled from different POVs from $(i-1)$-th scale.
  %  \item 
\end{itemize}

\subsubsection{NPAFormer - Transformer using Neighborhood Point Attention}
Earlier Transformers often conducted self-attention computation globally~\cite{Guo_2021}, inevitably incurring unbearable complexity for applications using large-scale point clouds. Recently, a number of explorations on 2D images~\cite{liu2021swin,lu2021transformer} proved that  examining self-attention in local neighborhood could retain almost the same performance in various tasks but with significant reduction of complexity. However, unlike 2D images with structured topology, point cloud is essentially an unordered set of points in 3D space that is difficult to model using a regular size window. This motivates us to develop the NPAFormer that constructs local neighborhood dynamically conditioned on individual POV input.% individual for better information aggregation.% $k$ nearest neighbors dynamically.

%Later Swin Transformer\cite{liu2021swin} perform self-attention in a local window to reduce complexity.Point cloud is essentially an unordered set of points in 3D space,it is natural to treat these points as tokens and aggregate features from points through attention mechanism.Recently published Point Transformer\cite{zhao2021point} and Point Cloud Transformer\cite{Guo_2021} have proved its potential in point cloud.However the quadratic complexity with respect to the number of tokens make self-attention hard to conduct on point cloud with a relative large amount of points.

As aforementioned, we embed NPAFormer in MOPA to best exploit the correlations from $k$ nearest neighbors for effective information aggregation. As seen in Fig.~\ref{sfig:mopa}, the structure of NPAFormer is similar to standard Transformer architecture, consisting a NPA layer for neighborhood point attention computation, normalization (Norm) and linear layers as well as the residual links~\cite{he2016deep}.

%For any POV in given sparse tensor input $\mathbb{S}_{in} = \{C_{in}\in\mathbb{R}^{n \times 3}, F_{in}\in\mathbb{R}^{N \times d_{e}}\}$,

%orhood Attention module,the illustration of this structure is presented in Figure\ref{}.
Assuming the input of a NPA layer is a sparse tensor consisting of $d_{in}$-dimensional features  $F_{in}\in\mathbb{R}^{N \times d_{in}}$ and $3$-dimensional coordinates  $C_{in}\in\mathbb{R}^{N \times 3}$. We then conduct the $k$NN search for each element in input sparse tensor, yielding dynamic $k$NN tensor $\left\{C_{\rm kNN}\in\mathbb{R}^{N \times k \times 3},F_{\rm kNN}\in\mathbb{R}^{N \times k \times d_{in}}\right\}$.

\textbf{Position Embedding.} Unconstrained displacement of points allows us to accurately and flexibly represent 3D space. But apparently it makes the embedding of geometry position more difficult by using the  absolute value. Instead, we propose to concatenate the relative position with corresponding features, e.g., 
\begin{equation}
    F_{e}=concat(F_{\rm kNN},C_{\rm kNN}-C_{in}), \mbox{~~} F_{e}\in\mathbb{R}^{N \times k \times (d_{in}+3)},
\end{equation} with which we can best retain the spatial coherency for information characterization in local neighborhood.

%Unlike image that all pixels are located in fixed grids and the relative position between different pixels are fixed,relative position of points in a point cloud is arbitrary continuous.Considering that the position information of points in a point cloud is rather important,we conduct position embedding by concatenating the $F_{knn}$ and relative position to the center point,the output of position embedding $F_{e}\in\mathbb{R}^{N \times k \times (d_{e}+3)}$ is computed as follows:
%\begin{equation}
 %   F_{e}=concat(F_{knn},C_{knn}-C)
%\end{equation}

\begin{wrapfigure}{r}{0.65\textwidth}
    \centering
    \includegraphics[width=0.63\textwidth]{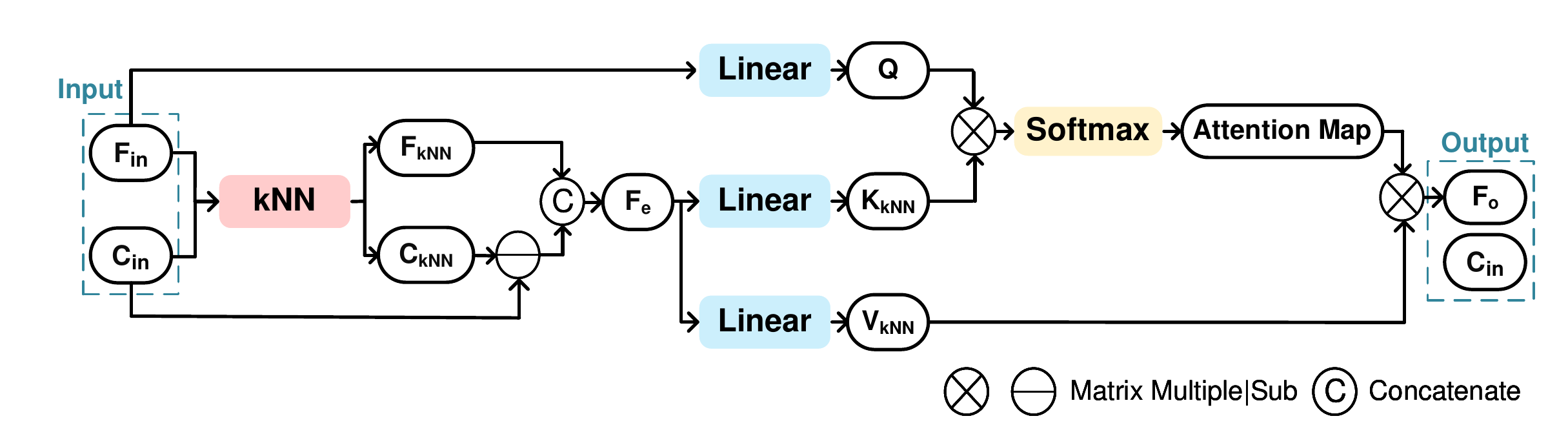}
    \caption{Neighborhood Point Attention Architecture. }
    \label{fig:npa}
\end{wrapfigure}
\textbf{Neighborhood Point Attention (NPA).} 
As revealed by its definition, the proposed NPA applies the attention based dynamic weighting among  $k$ nearest neighbors of each input POV (as the central point). %The mechanism of NPA can be simply described as allowing central point only attend to its neighbors.
The overall architecture of NPA is shown in Fig.~\ref{fig:npa}.

Let $Q$, $K_{\rm kNN}$ and $V_{\rm kNN}$ be the $query$, $key$, and $value$ vectors respectively. $Q \in \mathbb{R}^{N \times d_{e}}$ is computed through a linear transformation of $F_{in}$; $K_{\rm kNN}\in \mathbb{R}^{N \times k \times d_{e}}$ and $V_{\rm kNN}\in \mathbb{R}^{N \times k \times d_{e}}$ are computed by two separate linear transformations of $F_{e}$:
\begin{align}
    Q=F_{in}  \cdot W_{Q}, \mbox{~~} K_{\rm kNN} = F_{e} \cdot W_{K}, \mbox{~~} V_{\rm kNN} = F_{e} \cdot W_{V},
    %&Q \in \mathbb{R}^{N \times d_{e}} \quad K_{kNN},V_{kNN} \in \mathbb{R}^{N \times k \times d_{e}}\notag\\
\end{align} having linear transformations $ W_{Q} \in \mathbb{R}^{d_{in} \times d_{e}}$ and  $W_{K},W_{V} \in \mathbb{R}^{(d_{in}+3) \times d_{e}}$. Here,
$d_{in}$ is the dimension of input feature $F_{in}$ and $d_{e}$ is the dimension of $Q$, $K_{\rm kNN}$ and $V_{\rm kNN}$.  %$W_{Q},W_{K},W_{V}$ are the corresponding linear transformations.
Then the NPA can be calculated using:
\begin{equation}
    {\rm NPA}(Q,K_{\rm kNN},V_{\rm kNN})=softmax(\frac{Q K_{\rm kNN}^{T}}{\sqrt{d_{e}}})V_{\rm kNN}.
\end{equation}
The dot product of $Q$ and $K_{\rm kNN}$ is divided by $\sqrt{d_{e}}$ and then passed through the $softmax$ function to derive the Attention Map, then the dot product of Attention Map and $V_{\rm kNN}$ is denoted as $F_o$. The output of NPA comprises of $F_o$ and $C_{in}$.

\textbf{Multihead NPA.}
To best capture correlations from different representation subspaces, we extend aforementioned Singlehead NPA to Multihead NPA following the rules defined for standard Transformers~\cite{vaswani2017attention}.
We use several different linear projection upon the $query$, $key$, and $value$ vectors, and then feed them into multiple NPA module in parallel. The ouputs of parallel NPAs are concatenated together and passed through a linear transformation. We generally adapt  the number of attention heads (\#ah) in Multihead NPA, e.g., \#ah at $1,2\ldots4$, and enforce the product of  \#ah and associated channels per head \#cph fixed to 32 to have the fixed total dimensions for lightweight processing. As reported in~Table~\ref{tab:ablation_head}, the result shows that having \#ah = 4 and  \#cph = 8 is a balanced option.

% We have found that this small change produces the considerable improvement of compression performance compared with Singlehead NPA. One thing shall be marked is the fixed total dimensions used in this work for lightweight processing.%, e.g.,
%\begin{equation}
%$head_{i} \times d_{i} = head_{j} \times d_{j}$.    
%\end{equation}
%We enforce the product of  $head_i$ and associated channels per head $d_i$ fixed to 32. %, eliminating the possibility that the performance improvement comes from an increase of the number of channel.

%\textbf{Receptive Field.}
 %Previous works like {\it SparsePCGC}~\cite{wang2021sparse} build context model based on SConv performs well in object PCG whose spatial distribution is relatively dense but they are not efficient in LiDAR PCG. We comes down to the extremely sparse distribution of LiDAR PCG in 3D space and SparseCNN struggling to capture valid neighbor points. Our approach that search the neighbor points by $k$NN algorithm could further improve the expression ability of context model for LiDAR PCG.

\textbf{Complexity Analysis.}
Having NPA computation on local neighborhood with $k$ elements avoids the quadratic increase of computation and memory cost with respect to the number of points contrast with methods using Global Self-Attention. Here, for simplicity, we use the single-head NPA as an example for discussion. Multihead NPA can be easily extended. Assuming the total amount of points is $N$, the number of neighbors in local neighborhood is $k$, and the dimensional size of feature is $C$. Note that we often have $k<<N$ and $C<<N$ in practical settings (e.g., $k=16$, $C=32$ in our examples but $N$ is typically hundreds or tens of thousand).  Thus, the computation complexity of NPA and Global Self-Attention can be approximated as:
\begin{align}
    \Omega ({\rm GlobalSelfAttention})=3NC^{2}+2N^{2}C+N^{2} &\propto \mathcal{O}(N^2), \\
    \Omega ({\rm NPA})=NC^{2}+2NkC^{2}+8NkC+Nk &\propto \mathcal{O}(N).
\end{align} 
As we can see, our NPA greatly reduces the  complexity for attention computation, making our solution attractive in practices.

\section{Results}

\subsection{Dataset}
\label{subsect:dataset}
 Prevalent SemanticKITTI~\cite{Behley_2019_ICCV} and Ford sequences~\cite{mpegdataset} provided by the MPEG committee are evaluated.

\textbf{SemanticKITTI} is a publicly-accessible large-scale LiDAR point clouds dataset used for semantic scene understanding. It includes 43,552 raw scans (frames) with 4.5 billions points collected from a Velodyne HDL-64E sensor. We use sequences from \#00 to \#10 (23,201 scans in total) for training, and the remaining sequences from \#11 to \#21 (20,351 scans) for testing, following  the official training/testing split as suggested.

\textbf{Ford} is dynamically acquired for MPEG point clouds compression standardization. It contains 3 sequences, e.g., Ford$\_$01, Ford$\_$02 and Ford$\_$03, each of which includes 1500 frames at 1mm precision (or 18 bits per geometry component). We take all of them as our testing sequence.  

\subsection{Experiments}
\label{subsec:experiment}
\textbf{Baselines} include both rules- and learning-based PCG coding solutions for comparative studies: 
\begin{itemize}
    \item {\it G-PCC}~\cite{tmc13code} uses the latest reference model TMC13v14~\cite{tmc13code} with rules-based octree codec to generate anchor following the common test conditions (CTC) defined in~\cite{MPEG_GPCC_CTC}. 
    \item {\it SparsePCGC}~\cite{wang2021sparse} represents the use of multiscale sparse tensor but with stacked SConvs;
    \item {\it OctAttention}~\cite{fu2022octattention}, {\it VoxelContextNet}~\cite{Que_2021_CVPR} \& {\it OctSqueeze}~\cite{Huang_2020_CVPR} are learned octree codecs. 
\end{itemize}

We faithfully reproduce the results of {\it SparsePCGC} and {\it OctAttention} from their source codes. Since the source codes of {\it VoxelContextNet} and {\it OctSqueeze} are not publicly accessible, we directly cite the results from their papers.  Following the suggestions in {\it OctAttention}~\cite{fu2022octattention} for fair comparison, we remove the Coordinate Refinement Module (CRM) in both {\it VoxelContextNet} and {\it SparsePCGC} methods. Such CRM is a  post-processing module that can be augmented for all approaches.

%Then we set scaling factor "positionQuantizationScale" as $1$ for lossless compression.As for lossy compression,we set "positionQuantizationScale" from $\frac{1}{2}$ to $\frac{1}{256}$ in SematicKITTI test dataset and  $\frac{1}{2}$ to $\frac{1}{512}$ in Ford test dataset.

%Apart from that,we also conduct comparison with learning-based octree representation methods including state-of-the-art method OctAttention\cite{fu2022octattention},VoxelContext Net\cite{Que_2021_CVPR} and OctSqueeze\cite{Huang_2020_CVPR} in LiDAR PCG compression to prove the effectiveness of our method.Since the code of Voxel-Context Net\cite{Que_2021_CVPR} and OctSqueeze\cite{Huang_2020_CVPR} are not available,we directly cite the results in the paper.We test the OctAttention\cite{fu2022octattention} by ourselves because the code is available.For fairness,we compare with Voxel-Context Net\cite{Que_2021_CVPR} without Coordinate Refinement Module and SparsePCGC\cite{wang2021sparse} without Offset Module,which are more like a post-processing method.We follow the quantization settings of these methods to perform RD control.

\textbf{Training.}
% In the training procedure,limited by the memory size of gpu,we train two models with higher bit rates(scaling factor from $1$ to $\frac{1}{8}$) and lower bit rates(scaling factor from $\frac{1}{16}$ to $\frac{1}{128}$) respectively.  We set number of neighbors $k=16$ and $4$ heads Neighborhood Point Attention in NPAFormer when training the two models.
% In the testing procedure, we when comparing with SparsePCGC~\cite{wang2021sparse} and G-PCC. adjust scaling factor from $1$ to $\frac{1}{512}$ in Ford test dataset and $1$ to $\frac{1}{256}$ in SematicKITTI test dataset 
% set $L$ described in the paper\cite{fu2022octattention} from $8$ to $12$ to evaluate our model in different bit rates. 
% To conduct a more comprehensive experiment,since the code of OctAttention\cite{fu2022octattention} is available,we also test it and our method with larger $L$ from $13$ to $16$.
% To conduct a more comprehensive experiment,since the code of OctAttention\cite{fu2022octattention} is available,we also test it and our method with larger $L$ from $13$ to $16$.
% Our approach is implemented by PyTorch and MinkowskiEninge\cite{MinkowskiEngine},and we perform the training/testing on the computer with an Intel Xeon 6226R CPU and an Nvidia GeForce RTX 3090 GPU.
In the training procedure, we quantize native floating-point LiDAR geometry $P$ in SemanticKITTI training frames to 1mm unit precision by~\eqref{physicalquantization} first and use them to generate our training samples, e.g.,
\begin{equation}
    P_{Q}=round\left({P}/{precision}\right).
    \label{physicalquantization}
\end{equation} {As aforementioned, our MOPA works across the adjacent scales. Having the native resolution of training samples, e.g., scaling factor $S=1$, we downscale them to lower scales to form a set of pairs, e.g., (1/2, 1/4), (1/8, 1/16) etc, for model training.}

This work trains two models. One is for the relatively-low bitrate range mainly used by {\it OctAttention}, {\it VoxelContextNet} and {\it OctSqueeze}. To produces these low bitrates, input PCGs are downscaled with $S$ from $\frac{1}{16}$ to $\frac{1}{128}$. Note that the scaling factors can be mapped to the octree depth as in~\cite{fu2022octattention,Huang2020OctSqueezeOE,Que2021VoxelContextNetAO}. The other model is trained for high bitrates with $S$ from $1$ to $\frac{1}{8}$. Having an alternative high-bitrate model is because of the desire of (ultra) high-precision geometry precision in autonomous machinery where reconstruction of LiDAR PCGs at high bitrates is more preferable for application purpose. Finetuning the high-bitrate model is due to   very different characteristics (e.g., point density) at different scales. {Learned model is trained with binary cross entropy loss in supervised end-to-end means.}%On the other hand, as the desire of (ultra) high-precision geometry precision in autonomous machinery, reconstruction of LiDAR PCGs at high bitrates is more preferable for application purpose.

%generated by the dyadic downscaling factors from $\frac{1}{16}$ to $\frac{1}{128}$) and high bit rates (scaling factor from $1$ to $\frac{1}{8}$), respectively, to obtain optimal compression performance.

{\bf Testing} is enforced with the fair and best-reproducible principles in this study. We strictly follow the test conditions in other approaches or directly quote results from their publications.
Specifically, when comparing with the anchor like {\it G-PCC} or {\it SparsePCGC}~\cite{wang2021sparse}, we scale input LiDAR PCGs with $S$ from 1 to $\frac{1}{512}$ to obtain proper bitrates that match the anchor settings. % {\it G-PCC}, following the settings in {\it SparsePCGC}~\cite{wang2021sparse}.
 When comparing with prevalent learned octree coding approaches, e.g., 
{\it OctAttention}~\cite{fu2022octattention}, {\it VoxelContextNet}~\cite{Que_2021_CVPR}, and {\it OctSqueeze} ~\cite{Huang_2020_CVPR}, testing PCGs $P$ are quantized from 8 to 12 bits per geometry component through the use of ~\eqref{levelquantization} defined in~\cite{fu2022octattention}, e.g.,
\begin{align}
    P_{Q}=round({(P-offset)}/{q_s}), \quad q_s={2}/{(2^D -1)}, \quad offset=\min(P_x,P_y,P_z),
    \label{levelquantization}
\end{align}
$P$ is the normalized PCGs in $[-1,1]$, $D$ is the max depth of octree from 8 to 12. $P_x$, $P_y$ and $P_z$ represent coordinate level in a given PCG. 

%For a more complete comparison at high bitrates, we also faithfully retest {\it OctAttention}\cite{fu2022octattention} on $13$ to $16$ bits quantized point clouds with its open source code.

\textbf{Evaluation Metrics.}
% Another important metric is complexity,we quantitatively express it in codec time for reference only.
% compare performance where  the PSNRs (Peak signal-to-noise ratio) measured by both point to point error (D1) and point to plane error (D2), 
% We closely follow the CTC~\cite{MPEG_GPCC_CTC} to quantitatively compare performance where  the PSNRs (Peak signal-to-noise ratio) measured by both point to point error (D1) and point to plane error (D2), bpp as well as the BDBR~\cite{bjontegaard2001calculation} are used to report rate distortion efficiency.  PSNR and BDBR measurements for both D1 and D2 are calculated using tools recommended by the MPEG committee. Note that BDBR metric is used for lossy compression while the compress ratio gain calculated using bpp measures is for lossles mode.
% When comparing with G-PCC and SparsePCGC~\cite{wang2021sparse},we set PSNR peak
% value $r = 30000$. When comparing with OctAttention~\cite{fu2022octattention},Voxel-Context Net~\cite{Que_2021_CVPR} and OctSqueeze~\cite{Huang_2020_CVPR},since SemanticKITTI is normalized into $[-1,1]$,we set PSNR peak value $r = 1$ following their setting.
% We also include BDBR~\cite{bjontegaard2001calculation} as a metric when evaluating performance in lossy compression.Another important metric is complexity,we quantitatively express it in codec time for reference only.
We closely follow the G-PCC CTC~\cite{MPEG_GPCC_CTC} to measure the bit rate using bpp (bit per input point), and quantitatively measure the distortion using PSNR (Peak Signal-to-Noise Ratio) of both point to point error (D1) and point to plane error (D2). Note that  compression ratio gain measured by bpp  is for lossless mode while the BD-rate~\cite{bjontegaard2001calculation} evaluates the lossy compression. 

As different approaches may use different normalization for  evaluation. For fair comparison, we  replicate the settings used by them accordingly. For example, when comparing with {\it G-PCC} and {\it SparsePCGC}~\cite{wang2021sparse}, we set PSNR peak value to $30000$ following the G-PCC CTC~\cite{MPEG_GPCC_CTC}; When comparing with the {\it OctAttention}~\cite{fu2022octattention}, {\it VoxelContextNet}~\cite{Que_2021_CVPR} and {\it OctSqueeze}~\cite{Huang_2020_CVPR}, we normalize the points to the range of $(-1,1)$ and set PSNR peak value to 1 following their rules.

%We use BDBR (Bjontegaard Bit Rate)~\cite{bjontegaard2001calculation} to compute the relative bit rate savings for overall rate distortion efficiency comparison.

For complexity measurement, we report the encoding and decoding runtime for reference. Tests are examined on a computer with an Intel Xeon 6226R CPU and an Nvidia GeForce RTX 3090 GPU.

\begin{table}[t]
\centering
\caption{Compression Performance Evaluation Using LiDAR PCGs for both Lossless and Lossy Scenarios. Anchor is standardized {\it G-PCC} using TMC13v14, and {\it SparsePCGC} is also provided.}
\label{exp1-table}
\resizebox{\textwidth}{!}{%
\begin{tabular}{|c|cccc|cccc|}
\hline
\multirow{4}{*}{LiDAR PCGs} & \multicolumn{4}{c|}{\multirow{2}{*}{Lossless (bpp)}}                                                                                                                         & \multicolumn{4}{c|}{Lossy}                                                                                                                 \\ \cline{6-9} 
                             & \multicolumn{4}{c|}{}                                                                                                                                                       & \multicolumn{2}{c|}{D1-BD-rate}                                                   & \multicolumn{2}{c|}{D2-BD-rate}                              \\ \cline{2-9} 
                             & \multicolumn{1}{c|}{G-PCC}          & \multicolumn{1}{c|}{SparsePCGC~\cite{wang2021sparse}}     & \multicolumn{1}{c|}{Ours}           & \begin{tabular}[c]{@{}c@{}}Gain over\\ G-PCC\end{tabular} & \multicolumn{1}{c|}{SparsePCGC~\cite{wang2021sparse}}       & \multicolumn{1}{c|}{Ours}              & \multicolumn{1}{c|}{SparsePCGC~\cite{wang2021sparse}}       & Ours              \\ \hline
KITTI\_vox2cm                & \multicolumn{1}{c|}{7.62}           & \multicolumn{1}{c|}{7.06}           & \multicolumn{1}{c|}{6.18}           & -18.90\%                                                  & \multicolumn{1}{c|}{-10.22\%}         & \multicolumn{1}{c|}{-18.01\%}          & \multicolumn{1}{c|}{-8.40\%}          & -17.96\%          \\ \hline
Ford\_vox2cm                 & \multicolumn{1}{c|}{9.95}           & \multicolumn{1}{c|}{9.69}           & \multicolumn{1}{c|}{8.53}           & -14.27\%                                                  & \multicolumn{1}{c|}{-10.18\%}         & \multicolumn{1}{c|}{-18.53\%}          & \multicolumn{1}{c|}{-9.18\%}          & -18.53\%          \\ \hline
KITTI\_vox1mm                & \multicolumn{1}{c|}{20.17}          & \multicolumn{1}{c|}{19.73}          & \multicolumn{1}{c|}{16.62}          & -17.60\%                                                  & \multicolumn{1}{c|}{-6.80\%}          & \multicolumn{1}{c|}{-18.14\%}          & \multicolumn{1}{c|}{-6.24\%}          & -18.11\%          \\ \hline
Ford\_vox1mm                 & \multicolumn{1}{c|}{22.31}          & \multicolumn{1}{c|}{22.27}          & \multicolumn{1}{c|}{19.86}          & -10.98\%                                                  & \multicolumn{1}{c|}{-6.12\%}          & \multicolumn{1}{c|}{-16.08\%}          & \multicolumn{1}{c|}{-5.51\%}          & -16.07\%          \\ \hline
Average                      & \multicolumn{1}{c|}{\textbf{15.01}} & \multicolumn{1}{c|}{\textbf{14.69}} & \multicolumn{1}{c|}{\textbf{12.80}} & \textbf{-14.72\%}                                         & \multicolumn{1}{c|}{\textbf{-8.33\%}} & \multicolumn{1}{c|}{\textbf{-17.69\%}} & \multicolumn{1}{c|}{\textbf{-7.33\%}} & \textbf{-17.67\%} \\ \hline
\end{tabular}%
}
\end{table}

\begin{figure}
	\centering

		\includegraphics[scale=0.092]{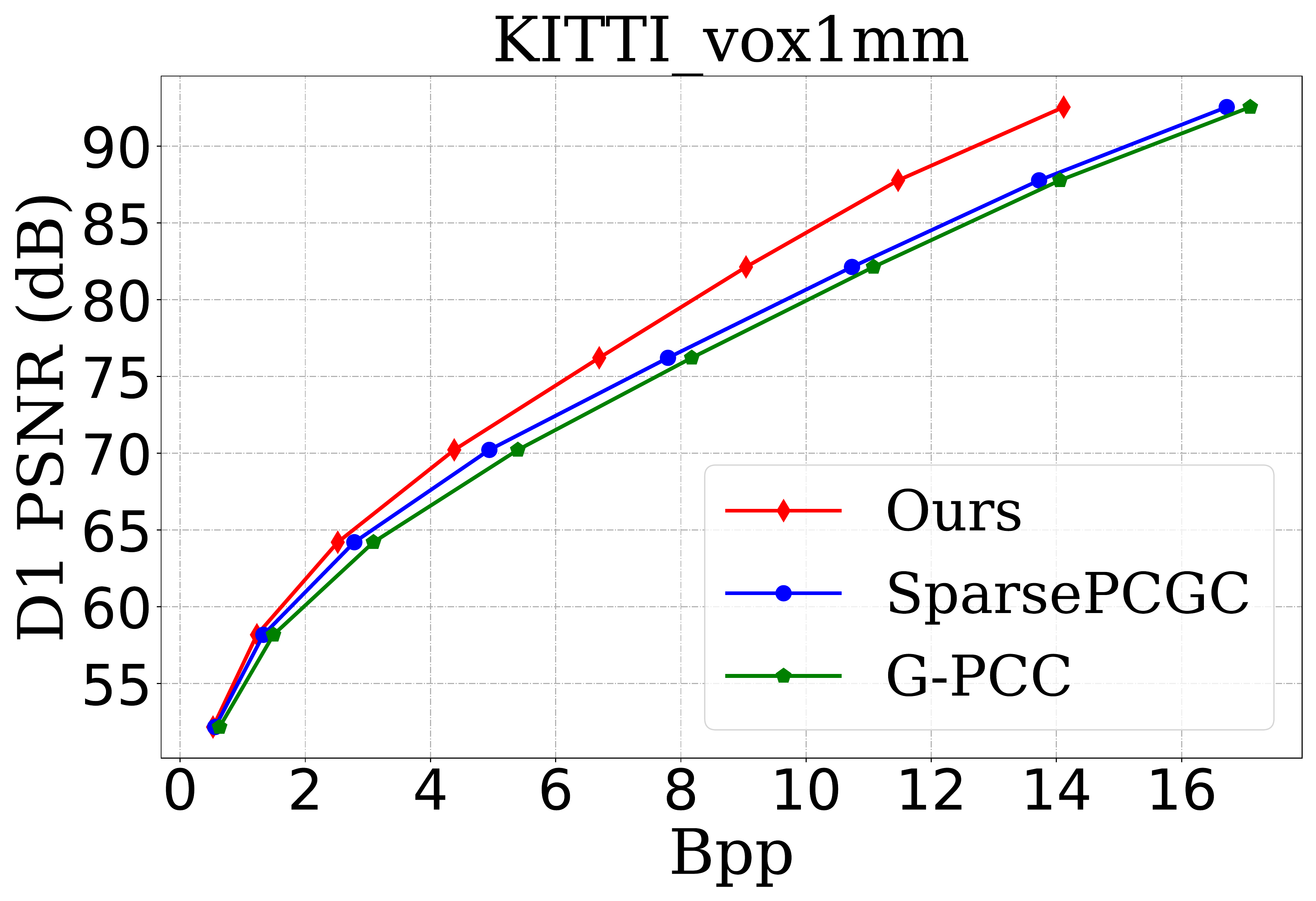}
		%}
%	\subfloat{
		\includegraphics[scale=0.092]{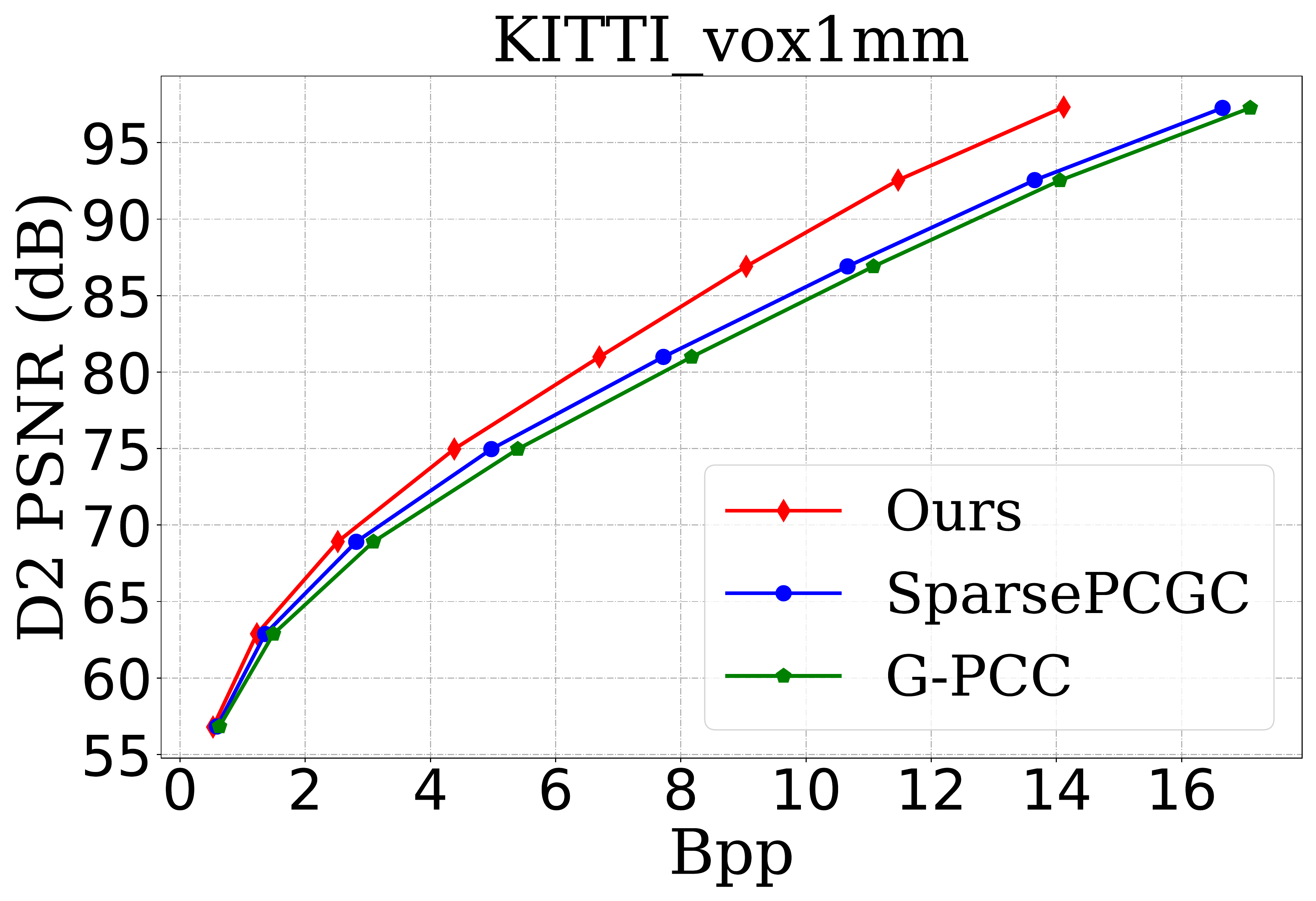}
		%}
%	\quad
%	\subfloat{
		\includegraphics[scale=0.092]{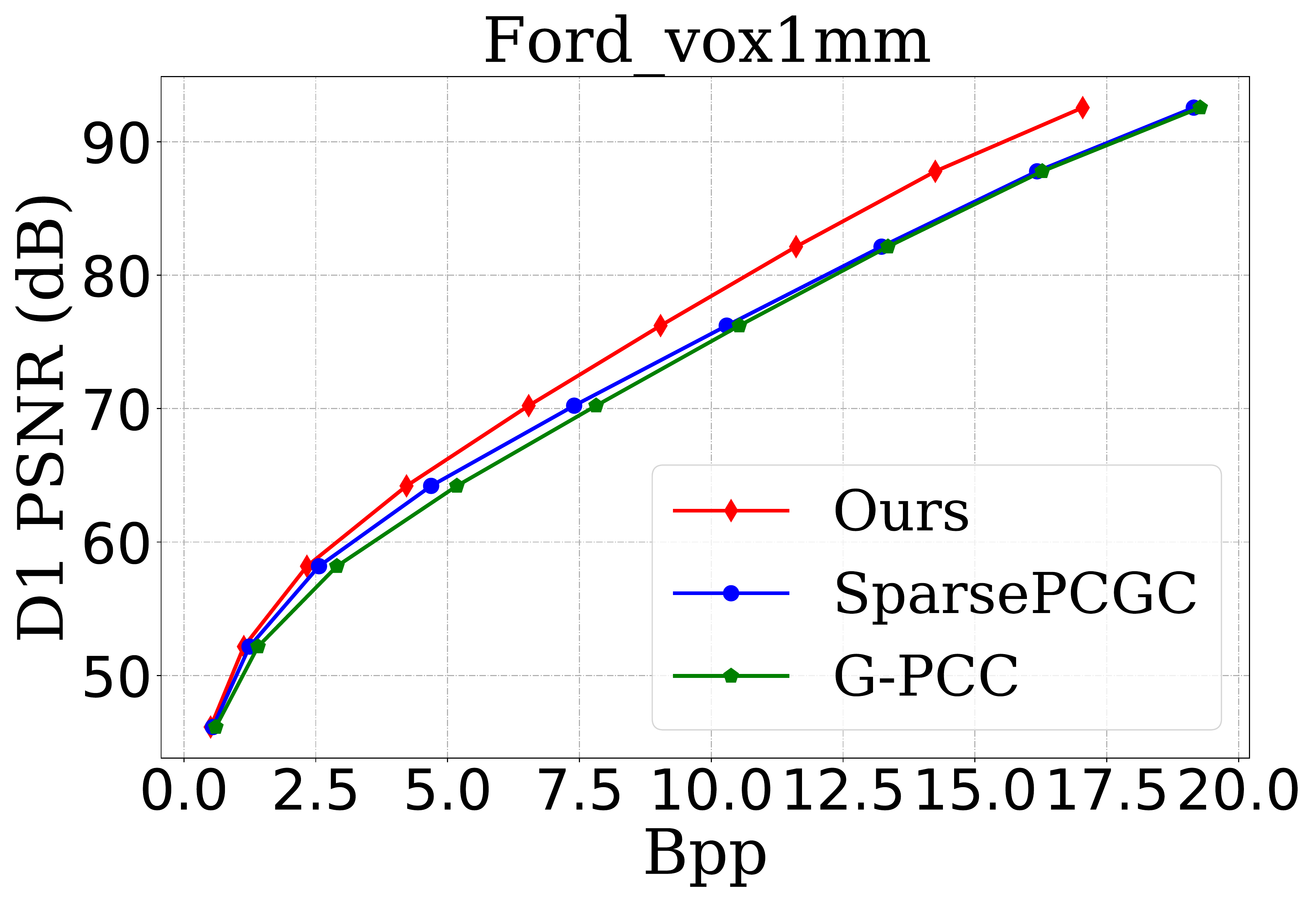}
		%}
%	\subfloat{
		\includegraphics[scale=0.092]{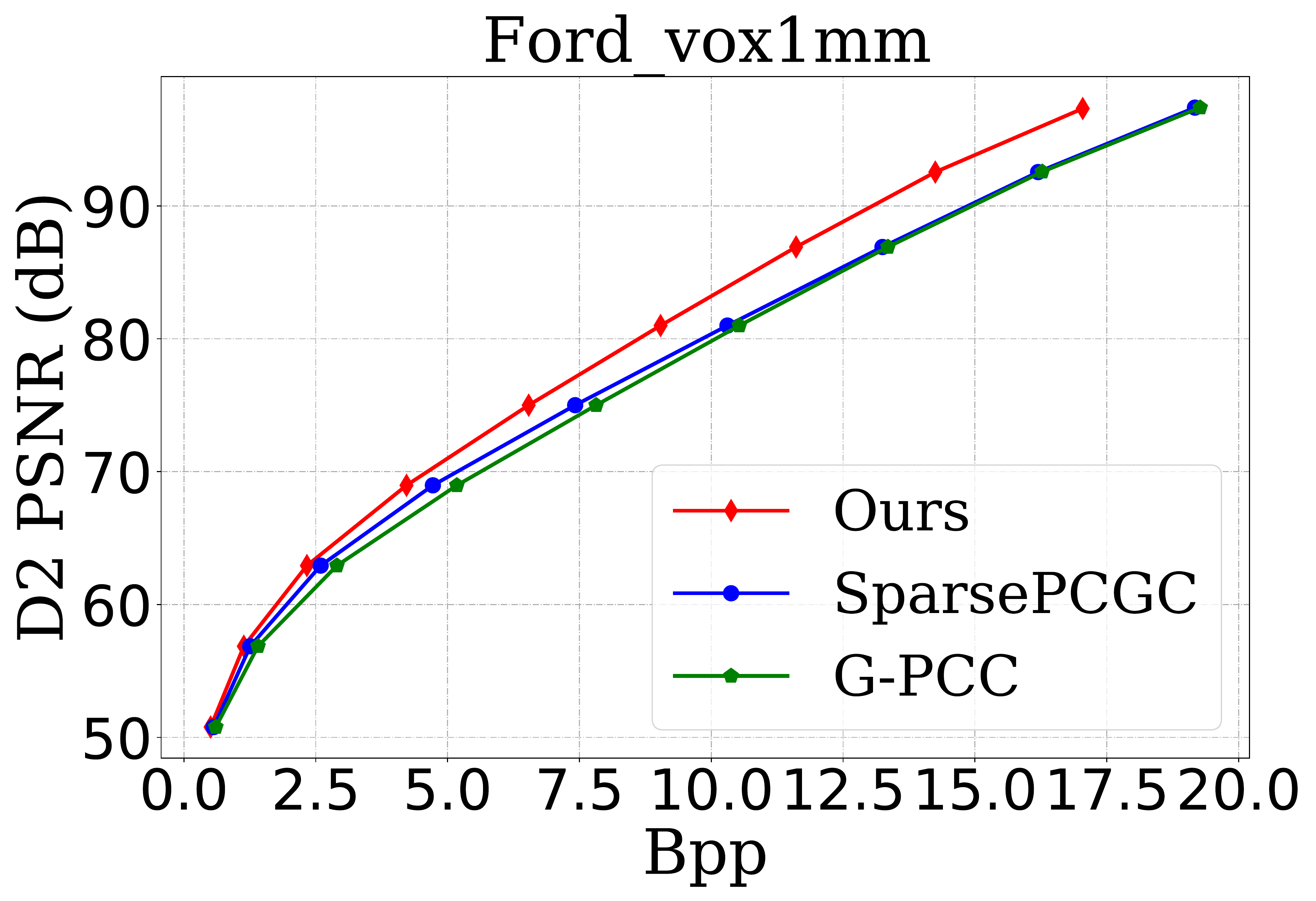}
		%}
	\caption{R-D comparison on SemanticKITTI and Ford samples across a wide range of bitrates.}
		\label{fig:lossy_r_d}
\end{figure}

\textbf{Quantitative Efficiency.} 
As shown in Table~\ref{exp1-table}, the proposed method offers more than 14\% and 17\% gains (on average) over the G-PCC anchor for respective lossless and lossy scenarios across a variety of test sequences at different precisions (1mm and 2cm). Performance improvements for lossy mode are also confirmed by sample rate-distortion (R-D) plots in Fig.~\ref{fig:lossy_r_d} where we can observe consistent improvement across a wide range of bitrates.

Having the same G-PCC anchor, our method offers $>$10 absolute percentage points increase for both lossy and lossless mode in comparison to the {\it SparsePCGC} (see Table~\ref{exp1-table} and Fig.~\ref{fig:lossy_r_d}). Given that our method uses the same multiscale sparse tensor as the {\it SparsePCGC}, the performance improvement reports the superior efficiency of NPA mechanism by resolving the limitations of stacked SConvs used in~\cite{wang2021sparse} to aggregate neighborhood information for better occupancy probability approximation. 

%The results show that our method outperforms baselines methods G-PCC and SparsePCGC~\cite{wang2021sparse} at all bitrates. On average, our method saves 14.72\% bitrates comparing with G-PCC and achieves 12.87\% bitrates saving against SparsePCGC~\cite{wang2021sparse} in lossless scene. Furthermore, our method bring about great improvement of saving 17.69\% and 17.67\% bitrates in terms of BDBR-D1 and BDBR-D2 respectively while SparsePCGC~\cite{wang2021sparse} saving only 8.33\% and 7.33\% corrspondingly in lossy scene. 

\begin{wrapfigure}{r}{0.5\textwidth}
\centering
\begin{minipage}[t]{0.49\textwidth}
	\centering
	%\subfloat{ 
		\includegraphics[scale=0.09]{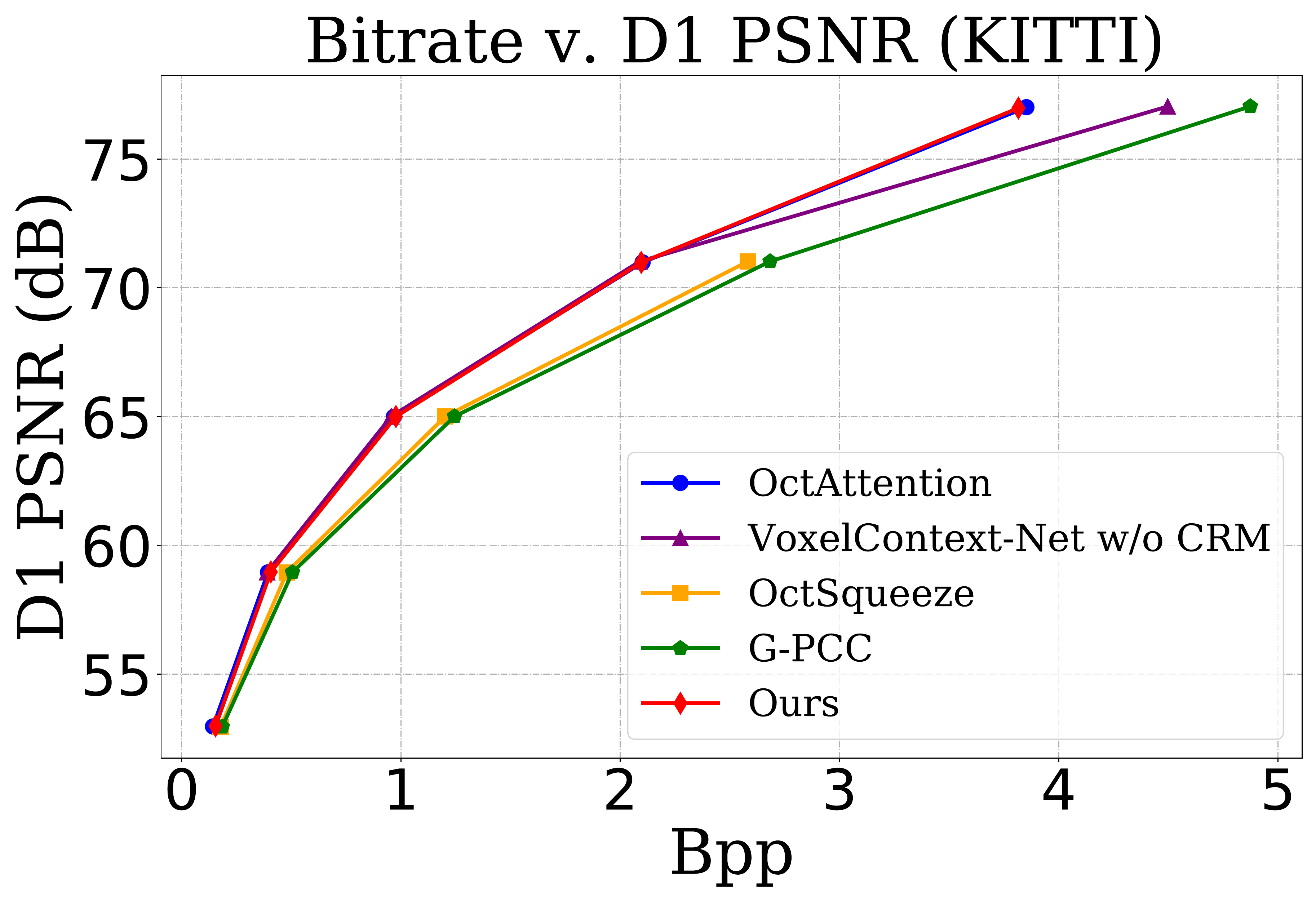}
		%}
	%\subfloat{
		\includegraphics[scale=0.09]{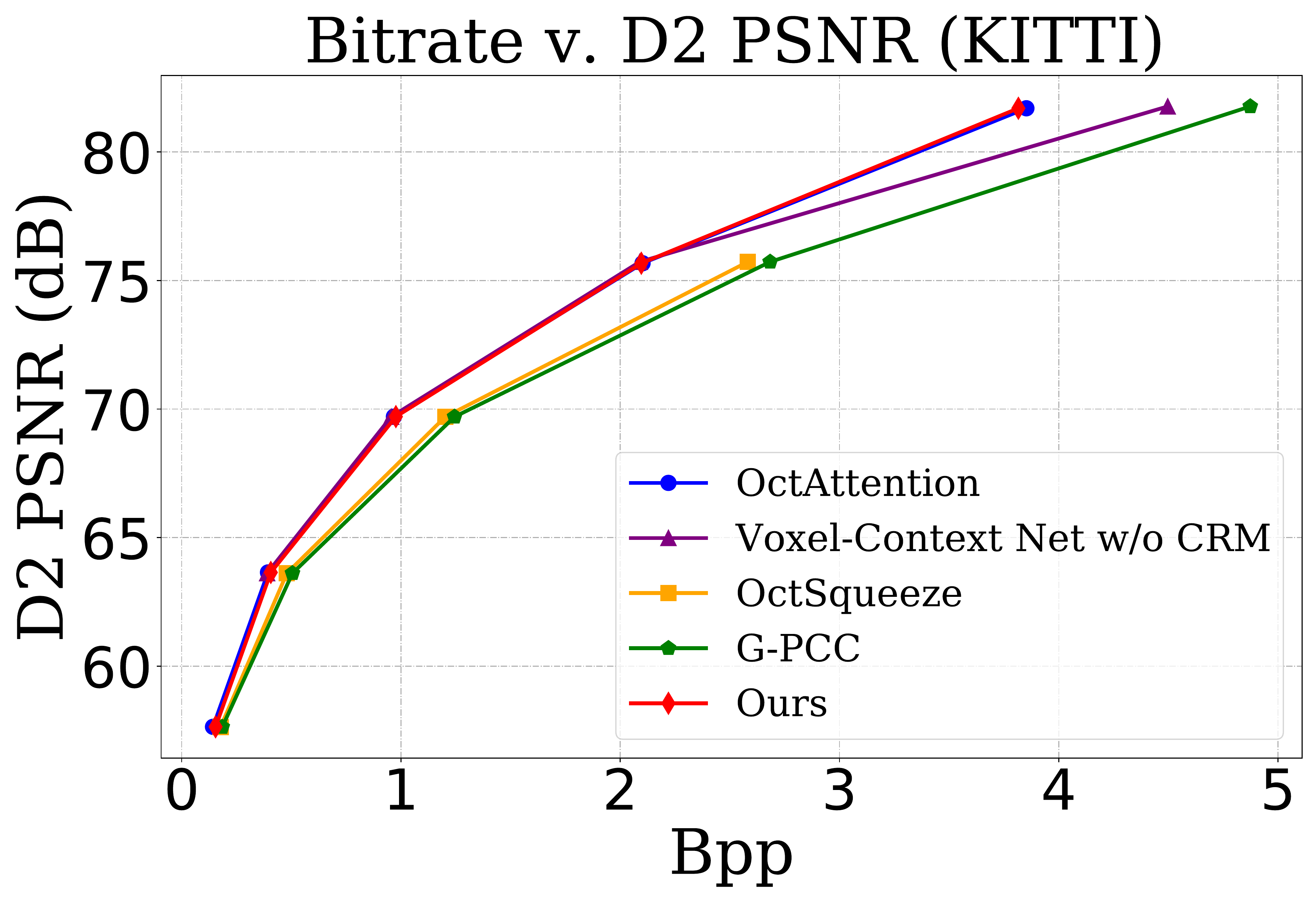}
		%}
	\caption{R-D comparison at lower bitrates}
		\label{fig:r_d_octattention_low}
\end{minipage}
\vfill
%\begin{table}[t]
\begin{minipage}[t]{0.49\textwidth}
	\centering
	%\subfloat{ 
		\includegraphics[scale=0.09]{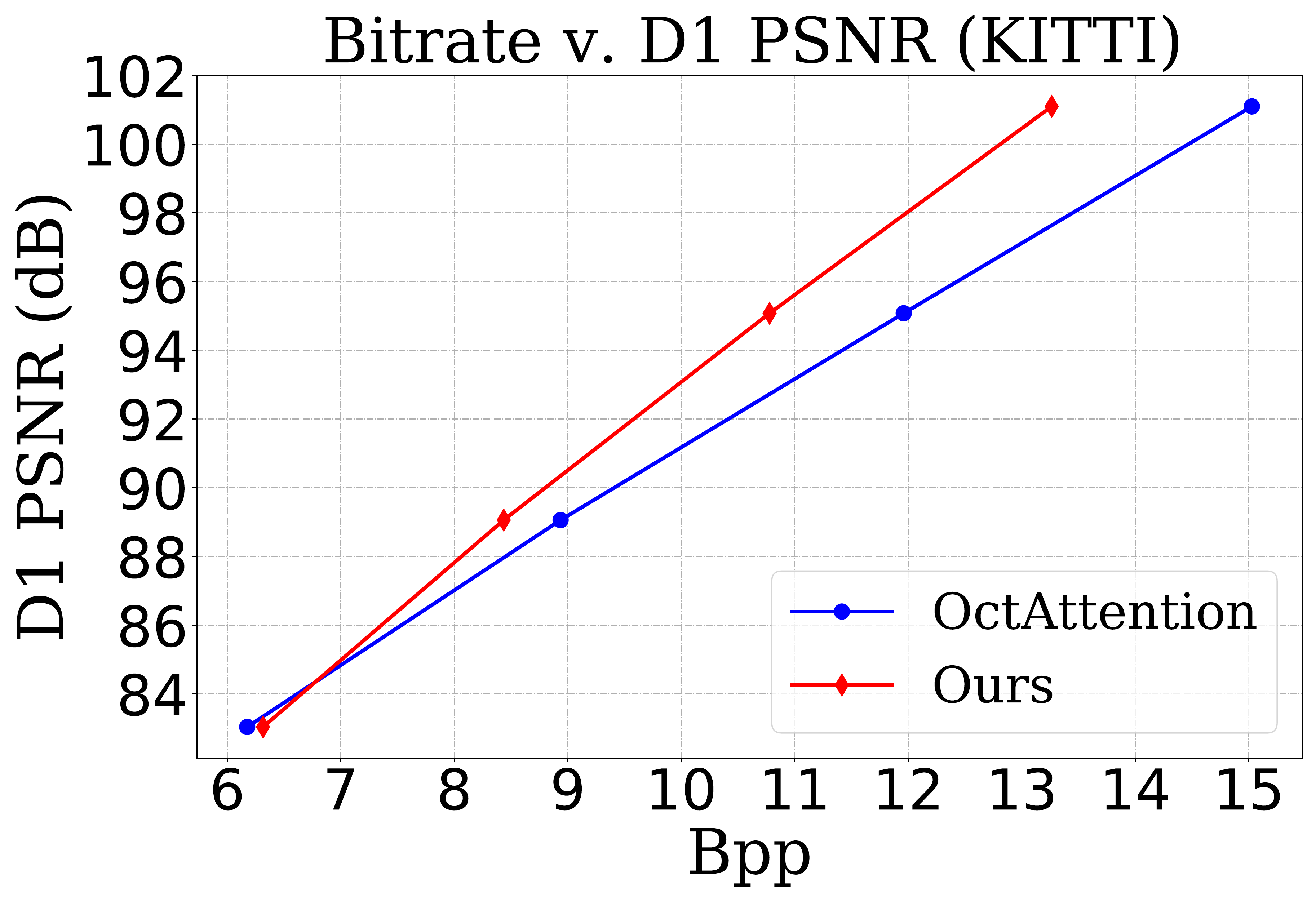}
		%}
	%\subfloat{
		\includegraphics[scale=0.09]{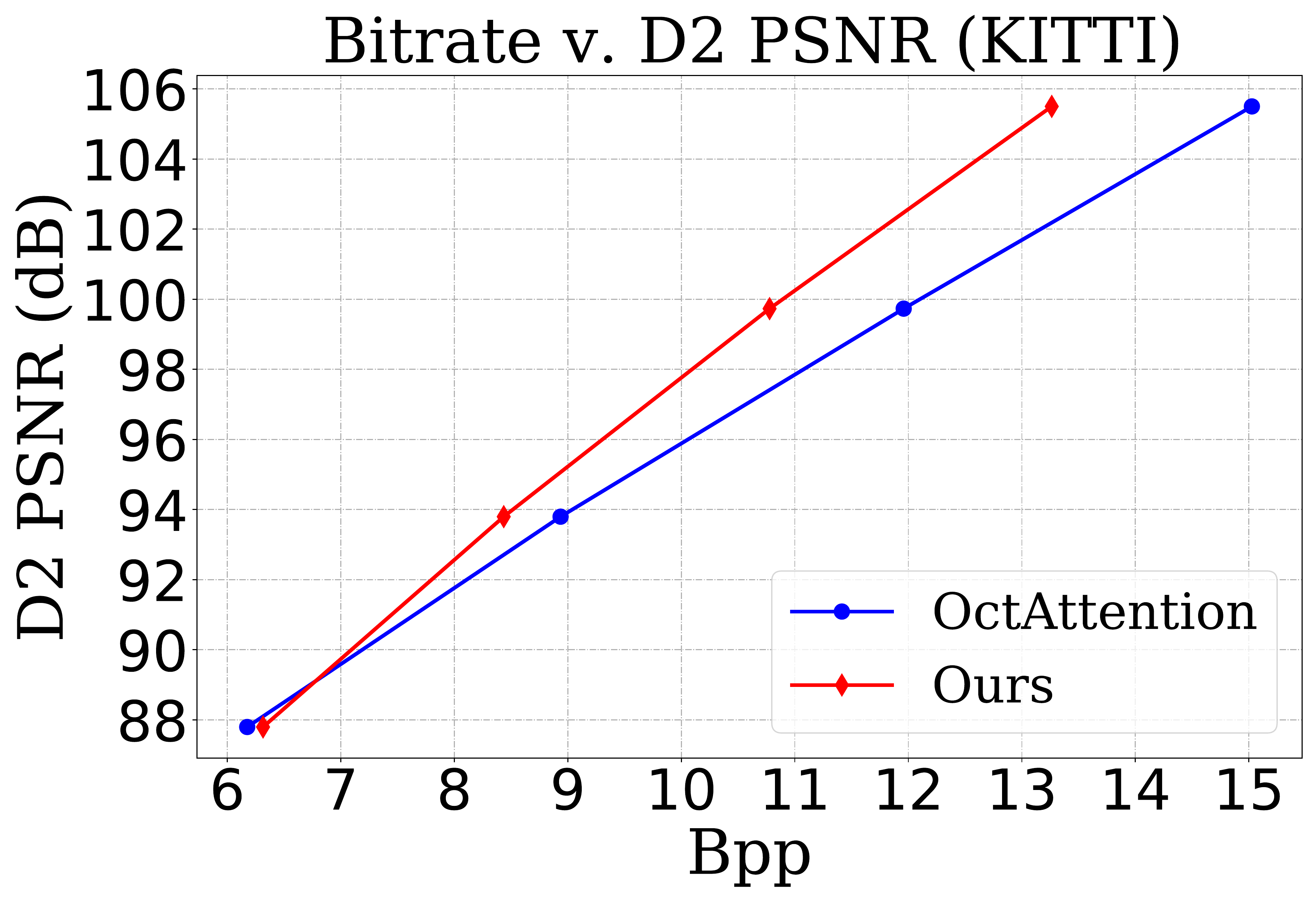}
		%}
	\caption{R-D comparison at higher bitrates}
		\label{fig:r_d_octtetntion_high}
\end{minipage}
%\end{table}
\end{wrapfigure}

%\textbf{Comparison with Learning-based Methods}
Comparisons are also conducted with other octree coding approaches as shown in Fig.~\ref{fig:r_d_octattention_low} where we apply the same bitrate range used by {\it OctAttention}, {\it VoxelContextNet} and {\it OctSqueeze}. These bitrates belong to a relatively lower range. Connecting with the experimental illustrations in SOTA {\it OctAttention}, our method shows almost the same R-D behavior with overlapped R-D curves in Fig.~\ref{fig:r_d_octattention_low}. As seen, our method and {\it OctAttention} report BD-rate gains over the {\it VoxelContextNet} at the point with the highest bpp, and consistently outperform the {\it OctSqueeze} and {\it G-PCC} for all bitrates.  

Our method shows even larger gains over the SOTA {\it OctAttention}, e.g., $\approx 7\%$ BD-rate improvement on average, when we extend the bitrates to higher level as shown in Fig.~\ref{fig:r_d_octtetntion_high}. This further evidences the superior efficiency of our method where the proposed NPA just using $k = 16$ local neighbors can offer better performance  than the default settings  in {\it OctAttention} with context window size at 1024 to gather the information of sibling and ancestor nodes as much as possible.

Note that we produce the high birate results of {\it OctAttention} by re-adapting its model to enable the max depth of octree from 13 to 16. {Unfortunately, we could not provide R-D performance at higher bitrates for {\it VoxelContextNet} and {\it OctSqueeze} because of the lack of their source codes for high bitrate model training.} However, since the {\it OctAttention} offers the SOTA efficiency, it is indeed a convincing representative of learned octree coding approaches.

Due to the page limitation, qualitative visualization is given in supplemental material where the improvement of the proposed method is clearly presented.

\begin{comment}
\begin{table}
\centering
\captionof{table}{Compression Efficiency}
\label{exp2-table}
\begin{tabular}{|c|cccc|}
\hline
\multirow{2}{*}{Methods} & \multicolumn{4}{c|}{Bpp on SemanticKITTI}                                                                                 \\ \cline{2-5} 
                        & \multicolumn{1}{c|}{D=13}  & \multicolumn{1}{c|}{D=14}           & \multicolumn{1}{c|}{D=15}            & D=16            \\ \hline
OctAttention\cite{fu2022octattention}            & \multicolumn{1}{c|}{6.176} & \multicolumn{1}{c|}{8.937}          & \multicolumn{1}{c|}{11.959}          & 15.027          \\ \hline
Ours                    & \multicolumn{1}{c|}{6.316} & \multicolumn{1}{c|}{\textbf{8.436}} & \multicolumn{1}{c|}{\textbf{10.778}} & \textbf{13.263} \\ \hline
\end{tabular}
\end{table}

%\end{minipage}

\begin{table}[]
\centering
\caption{tp}
\label{tab:tp}
\begin{tabular}{|cccc|}
\hline
\multicolumn{4}{|c|}{Compare with OctAttention}                                                                                     \\ \hline
\multicolumn{2}{|c|}{BDBR-D1}                                               & \multicolumn{2}{c|}{BDBR-D2}                          \\ \hline
\multicolumn{1}{|c|}{Lower bitrates} & \multicolumn{1}{c|}{Higher bitrates} & \multicolumn{1}{c|}{Lower bitrates} & Higher bitrates \\ \hline
\multicolumn{1}{|c|}{+1.66\%}        & \multicolumn{1}{c|}{-7.08\%}         & \multicolumn{1}{c|}{+1.75\%}        & -7.02\%         \\ \hline
\end{tabular}
\end{table}
\end{comment}

\begin{table}[b]
\caption{Comparisons of encoding and decoding time measured in seconds. Various octree depths $D$s are evaluated and SemanticKITTI sequences are exemplified. Time is measured for each sequence.}
\label{tab:encoding_decoding_time}
\begin{minipage}[t]{\textwidth}
        \centering
        \resizebox{0.85\textwidth}{!}{
        \begin{tabular}{ccccccccccc}
            \hline
            \multicolumn{11}{c}{Encoding Time}                                   \\ \hline
            \multicolumn{1}{c|}{Methods}      & \multicolumn{1}{c|}{D=8}  & \multicolumn{1}{c|}{D=9}  & \multicolumn{1}{c|}{D=10} & \multicolumn{1}{c|}{D=11} & \multicolumn{1}{c|}{D=12}  & \multicolumn{1}{c|}{D=13}  & \multicolumn{1}{c|}{D=14}  & \multicolumn{1}{c|}{D=15}  & \multicolumn{1}{c|}{D=16}  & Ave. 
            \\ \hline
            \multicolumn{1}{c|}{{\it G-PCC}}        & \multicolumn{1}{c|}{0.50} & \multicolumn{1}{c|}{0.62} & \multicolumn{1}{c|}{0.82} & \multicolumn{1}{c|}{1.07} & \multicolumn{1}{c|}{1.61}  & \multicolumn{1}{c|}{2.05}  & \multicolumn{1}{c|}{2.53}  & \multicolumn{1}{c|}{2.78}  & \multicolumn{1}{c|}{3.02}  & 1.67    
            \\ \hline
            \multicolumn{1}{c|}{{\it OctAttention}~\cite{fu2022octattention}} & \multicolumn{1}{c|}{0.40} & \multicolumn{1}{c|}{0.49} & \multicolumn{1}{c|}{0.54} & \multicolumn{1}{c|}{0.44} & \multicolumn{1}{c|}{0.66}  & \multicolumn{1}{c|}{1.51}  & \multicolumn{1}{c|}{1.91}  & \multicolumn{1}{c|}{2.22}  & \multicolumn{1}{c|}{2.58}  & 1.19    
            \\ \hline
            \multicolumn{1}{c|}{{\it SparsePCGC}~\cite{wang2021sparse}}   & \multicolumn{1}{c|}{2.08} & \multicolumn{1}{c|}{2.91} & \multicolumn{1}{c|}{4.36} & \multicolumn{1}{c|}{7.11} & \multicolumn{1}{c|}{10.58} & \multicolumn{1}{c|}{14.70} & \multicolumn{1}{c|}{18.68} & \multicolumn{1}{c|}{22.79} & \multicolumn{1}{c|}{20.88} & 11.57   
            \\ \hline
            \multicolumn{1}{c|}{Ours}         & \multicolumn{1}{c|}{1.18} & \multicolumn{1}{c|}{1.38} & \multicolumn{1}{c|}{1.79} & \multicolumn{1}{c|}{2.62} & \multicolumn{1}{c|}{4.11}  & \multicolumn{1}{c|}{5.35}  & \multicolumn{1}{c|}{8.05}  & \multicolumn{1}{c|}{10.65} & \multicolumn{1}{c|}{13.34} & 5.39    
            \\ \hline
        \end{tabular}
        }
        \label{exp3_enc_time}
\end{minipage}  
\\[5pt]  
\begin{minipage}[t]{\textwidth}
        \centering
        \resizebox{\textwidth}{!}{
        \begin{tabular}{ccccccccccc}
        \hline
        \multicolumn{11}{c}{Decoding Time}                                       \\ \hline
        \multicolumn{1}{c|}{Methods}      & \multicolumn{1}{c|}{D=8}   & \multicolumn{1}{c|}{D=9}    & \multicolumn{1}{c|}{D=10}   & \multicolumn{1}{c|}{D=11}   & \multicolumn{1}{c|}{D=12}    & \multicolumn{1}{c|}{D=13}  & \multicolumn{1}{c|}{D=14}  & \multicolumn{1}{c|}{D=15}  & \multicolumn{1}{c|}{D=16}  & Ave. 
        \\ \hline
        \multicolumn{1}{c|}{{\it G-PCC}}        & \multicolumn{1}{c|}{0.05} & \multicolumn{1}{c|}{0.09} & \multicolumn{1}{c|}{0.16} & \multicolumn{1}{c|}{0.31} & \multicolumn{1}{c|}{0.54}  & \multicolumn{1}{c|}{0.72}  & \multicolumn{1}{c|}{1.02}  & \multicolumn{1}{c|}{1.19}  & \multicolumn{1}{c|}{1.29}  & 0.60
        \\ \hline
        \multicolumn{1}{c|}{{\it OctAttention}~\cite{fu2022octattention}} & \multicolumn{1}{c|}{79.17} & \multicolumn{1}{c|}{227.79} & \multicolumn{1}{c|}{633.66} & \multicolumn{1}{c|}{782.11} & \multicolumn{1}{c|}{1569.79} & \multicolumn{1}{c|}{6066.57}      & \multicolumn{1}{c|}{6865.55}      & \multicolumn{1}{c|}{6628.52}      & \multicolumn{1}{c|}{8914.32}      &  3529.72       
        \\ \hline
        \multicolumn{1}{c|}{{\it SparsePCGC}~\cite{wang2021sparse}}   & \multicolumn{1}{c|}{2.88}  & \multicolumn{1}{c|}{3.54}   & \multicolumn{1}{c|}{4.76}   & \multicolumn{1}{c|}{6.97}   & \multicolumn{1}{c|}{11.19}   & \multicolumn{1}{c|}{14.51} & \multicolumn{1}{c|}{18.28} & \multicolumn{1}{c|}{22.43} & \multicolumn{1}{c|}{21.39} & 11.77   
        \\ \hline
        \multicolumn{1}{c|}{Ours}         & \multicolumn{1}{c|}{1.56}  & \multicolumn{1}{c|}{1.89}   & \multicolumn{1}{c|}{2.21}   & \multicolumn{1}{c|}{2.93}   & \multicolumn{1}{c|}{3.85}    & \multicolumn{1}{c|}{5.87}  & \multicolumn{1}{c|}{7.47}  & \multicolumn{1}{c|}{11.68} & \multicolumn{1}{c|}{12.48} & 5.55    
        \\ \hline
        \end{tabular}
        }    
        \label{exp3_dec_time}  
\end{minipage}
\label{exp3_enc_dec_time}
\end{table}

\begin{comment}
\begin{table}[]
\centering
\caption{Comparison of Model Size measured in MBytes.}
\label{tab:model_size}
\begin{tabular}{|c|c|c|c|}
\hline
Methods    & {\it OctAttention}~\cite{fu2022octattention} & {\it SparsePCGC}~\cite{wang2021sparse} & Ours  \\ \hline
Model Size & 16.13M       & 3.95M      & 8.96M \\ \hline
\end{tabular}
\end{table}
\end{comment}

\begin{table}[t]
\begin{minipage}[t]{0.27\textwidth}
\centering
\begin{scriptsize}
\captionof{table}{Model sizes}
\label{tab:model_size}
\begin{tabular}{c|c}
\hline
Method & Mbytes\\
\hline
{\it OctAttention}~\cite{fu2022octattention}   & 16.13 \\
{\it SparsePCGC}~\cite{wang2021sparse} & 3.95\\
Ours & 8.96 \\
\hline
\end{tabular}
\end{scriptsize}
\end{minipage}
\hfill
\begin{minipage}[t]{0.69\textwidth}
\centering
\caption{Impact of num. of attention heads on Bpp at various $S$s.}
\label{tab:ablation_head}
\resizebox{\textwidth}{!}{%
\begin{tabular}{|c|ccccccccc|}
\hline
{\#ah} &  \multicolumn{1}{c|}{S=1/256}         & \multicolumn{1}{c|}{S=1/128}         & \multicolumn{1}{c|}{S=1/64}          & \multicolumn{1}{c|}{S=1/32}          & \multicolumn{1}{c|}{S=1/16}          & \multicolumn{1}{c|}{S=1/8}           & \multicolumn{1}{c|}{S=1/4}            & \multicolumn{1}{c|}{S=1/2}            & S=1            \\ \hline
1                      & \multicolumn{1}{c|}{\textbf{0.52}}          & \multicolumn{1}{c|}{\textbf{1.21}}          & \multicolumn{1}{c|}{\textbf{2.48}}          & \multicolumn{1}{c|}{\textbf{4.34}}          & \multicolumn{1}{c|}{6.71}          & \multicolumn{1}{c|}{9.23}          & \multicolumn{1}{c|}{11.92}          & \multicolumn{1}{c|}{14.74}          & 17.57          \\ \hline
2                      & \multicolumn{1}{c|}{\textbf{0.52}}          & \multicolumn{1}{c|}{1.22}          & \multicolumn{1}{c|}{2.51}          & \multicolumn{1}{c|}{4.37}          & \multicolumn{1}{c|}{\textbf{6.68}}          & \multicolumn{1}{c|}{9.06}          & \multicolumn{1}{c|}{11.55}          & \multicolumn{1}{c|}{14.24}          & 16.78          \\ \hline
4                      & \multicolumn{1}{c|}{0.53} & \multicolumn{1}{c|}{1.23} & \multicolumn{1}{c|}{2.52} & \multicolumn{1}{c|}{4.38} & \multicolumn{1}{c|}{6.70} & \multicolumn{1}{c|}{\textbf{9.04}} & \multicolumn{1}{c|}{\textbf{11.47}} & \multicolumn{1}{c|}{\textbf{14.12}} & \textbf{16.62} \\ \hline
\end{tabular}%
}
\end{minipage}
\end{table}

\textbf{Complexity} is critical for the application of LiDAR PCG compression in real-life scenarios. Comparisons are performed among representative methods including the {\it G-PCC} using rules-based octree coding, {\it OctAttention} using learned octree coding, {\it SparsePCGC} using multiscale sparse tensor with learned SConvs, and our method.  We first compare the encoding and decoding time respectively at different bitrates with results shown in Table~\ref{exp3_enc_dec_time}.  As seen, conventional {\it G-PCC}  demonstrates lightweight and balanced complexity consumption for encoding and decoding.  Note that {\it G-PCC} runs mainly on CPU while other learned methods heavily rely on the GPU and CPU. Thus the measures of {\it G-PCC} are just used as the reference to understand the computational complexity in general.% in a relative means. 

Though {\it OctAttention} presents even less encoding time (on average) than the {\it G-PCC}, its decoding complexity is unbearable. This is because at encoder, massive parallelism can be fulfilled due to the full availability of information, while in decoder, causal dependency strictly limits the sequential processing. 

Our method shows comparable complexity for encoding and decoding, e.g., less than 6 seconds on average. It is about 640$\times$ speedup of decoding when compared with the {\it OctAttention}, and about 2$\times$ speedup of {\it SparsePCGC}. It is worth to point out that the runtime of our method can be further greatly reduced, {since our current implementation is just an evaluation prototype. For instance, $k$NN search is not optimized, requiring frequent interruptions for data transfer between GPU and CPU, which apparently increases the processing time. A speedup of the encoding and decoding by another 5$\times$ is possible according to our pilot exploration.} 

Model size measures the space complexity. Table~\ref{tab:model_size} lists the consumption in Mbytes for {\it OctAttention}, {\it SparePCGC} and ours. As seen, our method requires about a half of model size to the {\it OctAttention}, while it doubles the size as compared to that of the {\it SparsePCGC}. 
{Lately, we have noticed that the increase of model size of the proposed method mainly comes from the ResNet modules in NPAFormer which can be removed without noticeable performance loss according to our preliminary study. This is deferred as our future work. }

\begin{table}[t]
\centering
\caption{Impact of number of nearest neighbors $k$ on Bpp at various $S$s}
\label{tab:ablation_neighborhood}
\resizebox{0.69\textwidth}{!}{%
\begin{tabular}{|c|c|c|c|c|c|c|c|c|c|}
\hline
k  & S=1/256       & S=1/128       & S=1/64        & S=1/32        & S=1/16        & S=1/8         & S=1/4          & S=1/2          & S=1            \\ \hline
8  & 0.54          & 1.25          & 2.59          & 4.56          & 6.95          & 9.36          & 11.85          & 14.55          & 17.05          \\ \hline
16 & \textbf{0.53} & \textbf{1.23} & \textbf{2.52} & \textbf{4.38} & \textbf{6.70} & \textbf{9.04} & \textbf{11.47} & \textbf{14.12} & \textbf{16.62} \\ \hline
24 & \textbf{0.53} & \textbf{1.23} & 2.53          & 4.41          & 6.73          & 9.09          & 11.56          & 14.24          & 16.67          \\ \hline
32 & \textbf{0.53} & 1.24          & 2.55          & 4.45          & 6.76          & 9.14          & 11.63          & 14.32          & 16.89          \\ \hline
\end{tabular}%
}
\end{table}

\subsection{Ablation Studies}
This section examines the impact of neighborhood size $k$ and number of attention heads \#ah to get more insights on the proposed NPA. Extra examinations are given in supplemental material.

\textbf{Number of Attention Heads \#ah.} 
% 5.41% reduction.S=1 lossless
% 2.64% saving bitrates in terms of BDBR-D1 at higher bitrates S=1/16-1/2
% 1.47% more bitrates in terms of BDBR-D1 at lower bitrates S=1/16-1/2
We adapt \#ah  at different bitrates to understand its impact on compression efficiency.
To this aim, we adjust the scaling factor $S$ from 1 (lossless) to 1/256 to reach different bitrate points. Since the distortion is the same for different \#ah at the same $S$, we only present the bpp in Table~\ref{tab:ablation_head}. to report the performance, i.e., the smaller bpp the better compression efficiency. In the end, we choose \#ah = 4 in this work. Although the setting of \#ah = 4 provides marginal loss at lower bitrates (e.g., $\approx1\%$ BD-rate loss when $S$ is from 1/256 to 1/32), it improves the coding efficiency noticeably at higher bitrates, e.g., 5.41\% bpp reduction when $S$ = 1 (lossless mode), and $>$ 2.5\% BD-rate improvement when $S$ is from  1/16 to 1/2.

%To comprehensively utilize various aspects of information in LiDAR point cloud, we adopt Multihead NPAFormer architecture. In this experiment, we alter the number of attention heads to observe the impact it brings about. We evaluate the model with different number of attention heads at various bitrates on KITTI\underline \ vox1mm  and the result is shown in Table~\ref{tab:ablation_head}. Note that the number of parameters in models with various number of attention heads is consistent. We find that the model with more attention head tends to have a great impact on higher bitrates and similar impact on lower bitrates. Specifically, the model with 4 attention heads achieved 5.41\% reduction of Bpp in lossless scene when $S=1$ and 2.64\% bitrates saving in terms of BDBR-D1 when $S$ is from $\frac{1}{16}$ to $\frac{1}{2}$ in lossy scene, but 1.47\% bitrates more in terms of BDBR-D1 when S is from $\frac{1}{256}$ to $\frac{1}{32}$.  

\textbf{Number of Nearest Neighbors $k$.} Table~\ref{tab:ablation_neighborhood} gives averaged Bpps at different $S$s when adapting the $k$ in proposed NPAFormer for attention. Similarly, the smaller Bpp the better coding efficiency (for the same distortion at a given $S$).
As seen, $k$ = 16 gives the best compression performance.
%As seen, $k$ = 16 gives the best trade-off of compression performance and complexity. 
Having a $k$ smaller than 16, performance is deteriorated because of insufficient neighbors used for information aggregation; while having a larger $k$ may include more irrelevant (or uncorrelated) neighbors, making the attentive weighting compromised. Also, larger $k$ expects extra computation overhead potentially. Recalling the comparison with the {\it OctAttention} in Fig.~\ref{fig:r_d_octattention_low} and~\ref{fig:r_d_octtetntion_high}, a small number of local neighbors may be already sufficient for constructing accurate contexts used in compression.

%Recalling the definition of $k$ in previous discussions,  The number of neighborhood size is directly related to the size of receptive field in our proposed NPAFormer, we are curious about how much our context model can be influenced by expanding neighborhood size $k$. However, an excessively large number of neighborhood size would bring about a large amount of time and memory consumption. To trade off compression performance and consumption, we conduct this experiment and the result is shown in Table~\ref{tab:ablation_neighborhood}. To our surprise, a large size of neighborhood for our compression model does not seem necessarily to be beneficial. The NPAFormer with $k$=16 performs better and needs less time consumption. Thereforewe we set $k$=16.

\section{Discussion}

%\section{Acknowledgement}
%Our sincere gratitude goes to the authors of

Efficient compression of LiDAR point clouds is increasingly demanded for the enabling of networked autonomous machinery where high-precision LiDAR PCGs can be shared for various tasks. However, the inter-correlation among dynamically acquired and unevenly distributed sparse LiDAR points is hard to characterize by convolutions with fixed receptive field and fixed weights (after training). This work therefore suggested the neighborhood point attention (NPA) to adaptively weigh the contributions from $k$ nearest neighbors that are constructed dynamically conditioned on the input point for effective information aggregation and embedding. We devised this NPA together with the multiscale sparse tensor representation, showing the SOTA efficiency on the compression of typical LiDAR PCGs in both lossy and lossless modes. More importantly, the proposed method also demonstrated encouraging complexity-performance tradeoff even under current prototype implementation, e.g, few seconds for both encoding and decoding as exemplified. 

\section{Supplemental Material}
This companion document supplements the main text with additional experiments and discussions, further evidencing the superior efficiency and robust generalization of the proposed method.

\subsection{Qualitative Visualization}
\begin{figure}[h]
	\centering
	\includegraphics[width=1\textwidth]{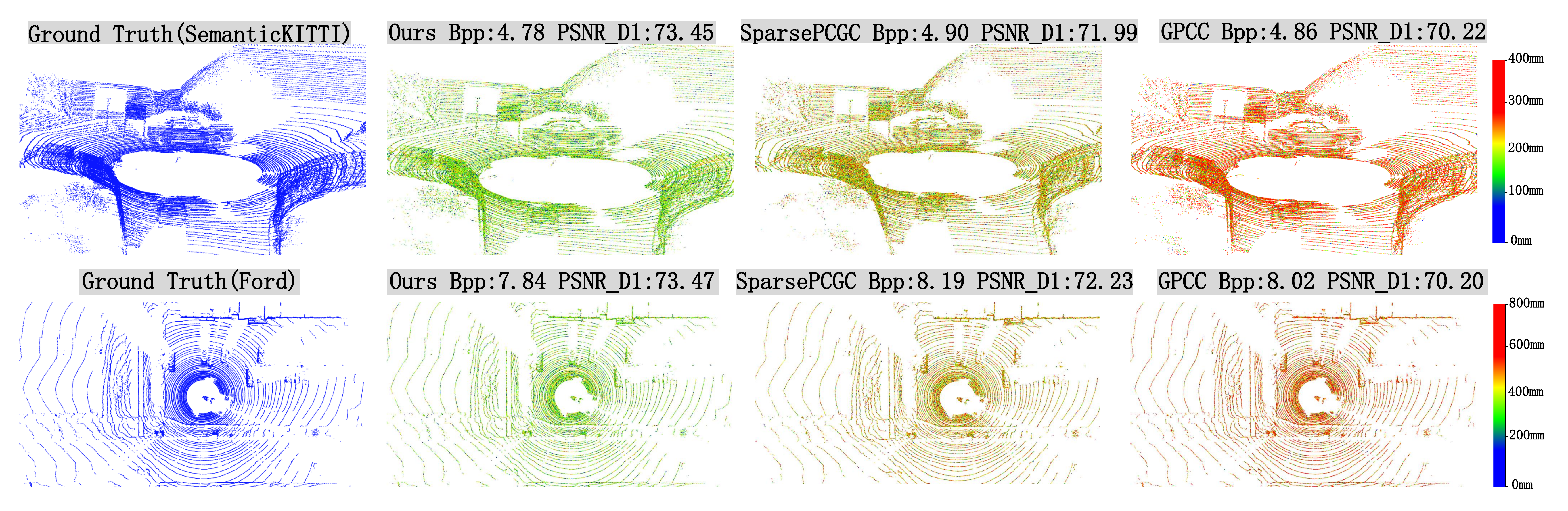}
	\caption{Qualitative visualization of LiDAR PCG reconstructions at typical bit rates. Color map is placed  to reflect the reconstruction error distribution. }
	\label{fig:qualitative_visualization_rec}
\end{figure}

We mainly give the quantitative evaluations in the main text (see Sec. 4.2) due to the page limitation. Here, we offer additional qualitative visualizations in Fig.~\ref{fig:qualitative_visualization_rec} to further demonstrate the efficiency of our method. As we can clearly observe, our method provides the least reconstruction error at close bitrates for both SemanticKITTI and Ford test samples, e.g., as for the \#000000 frame in SemanticKITTI \#11 sequence, the Bpp of our method is 4.78 and the corresponding PSNR of the point-to-point (D1) error is 73.45 dB, while {\it SparsePCGC}~\cite{wang2021sparse} and {\it G-PCC} present lower PSNR at respective 71.99 dB \& 70.22 dB and cost more bits, e.g., 4.90 Bpp \& 4.86 Bpp. The similar observation happens for other testing frames.      

\subsection{Model Performance on Object Point Clouds}
Here we report the performance using extra object point clouds that are recommended by MPEG standardization committee to present the efficiency evaluations as well as the robust generalization of the proposed method. We divide the experiment into two parts, one is for the solid object point clouds and the other is for the dense object point cloud. The difference between this two types of point clouds is that solid object point clouds have a higher density and lower precision compared with dense object point clouds.

\subsubsection{Solid Object Point Clouds}
In this subsection, we evaluate our method using solid object point clouds whose geometry distribution characteristic is totally different from the LiDAR PCGs studied in the main text. Baseline methods consist of MPEG {\it G-PCC}, {\it SparsePCGC}~\cite{wang2021sparse} and {\it OctAttention}~\cite{fu2022octattention}. The anchor result of {\it G-PCC} is generated using the latest TMC13v14~\cite{tmc13code}, and we reproduce the result of {\it SparsePCGC}~\cite{wang2021sparse} and {\it OctAttention}~\cite{fu2022octattention} faithfully following their methods described in the papers.

\textbf{Dataset.} Following the setting of {\it OctAttention}~\cite{fu2022octattention}, we use Soldier10 and Longdress10 sequences in MPEG 8i~\cite{8i20178i}, Andrew10, David10, and Sarah10 sequences in MVUB~\cite{microsoft2019microsoft} for training. Other point clouds in MPEG 8i~\cite{8i20178i} including Thaidancer, Boxer, Redandblack and Loot are selected for testing. Note that the PCGs mentioned above are presented with 9 or 10 bit precision. 

\textbf{Training and Testing.} In the training procedure, we train one single model with scaling factor $S$ from 1 to $\frac{1}{8}$. In the testing phase, being aligned with the {\it OctAttention}~\cite{fu2022octattention} that merely lossless compression results are offered, we only evaluate and compare the performance  in lossless mode. Other settings including evaluate metrics and loss function are the same as described in Sec. 4.2.

\textbf{Results.} The result of lossless compression for solid object PCGs is detailed in Table~\ref{tab:lossless_9_10_pcg}. It shows that our method achieves superior performance compared with previous methods including the {\it G-PCC}, {\it SparsePCGC}~\cite{wang2021sparse} and {\it OctAttention}~\cite{fu2022octattention}. Note that the number of neighbors $k$ is increased to 64 to have more neighbors for information aggregation, and the performance improvement shown in Table~\ref{tab:lossless_9_10_pcg} has proved its efficiency compared with $k$ = 16 and $k$ = 32. Even though, $k$ = 64 is still much smaller than the 1024 elements used in OctAttention. Our method achieves 37\% and 40\% gains over G-PCC on average in terms of compression ratio measured by Bpp for lossless compression.

\begin{table}[]
	\centering
	\caption{Lossless Compression Gains for Solid Object PCGs Measured by Bpp}
	\label{tab:lossless_9_10_pcg}
	\resizebox{1\textwidth}{!}{%
		\begin{tabular}{cc|c|cc|cc|cccccc}
			\hline
			\multicolumn{2}{c|}{\multirow{3}{*}{Solid Object PCGs}}                       & G-PCC                & \multicolumn{2}{c|}{SparsePCGC~\cite{wang2021sparse}}                                                                                        & \multicolumn{2}{c|}{OctAttention~\cite{fu2022octattention}}                                                                                      & \multicolumn{6}{c}{Ours}                                                                                                                                                                                                                                                                                                          \\ \cline{3-13} 
			\multicolumn{2}{c|}{}                                                         & \multirow{2}{*}{Bpp} & \multicolumn{1}{c|}{\multirow{2}{*}{Bpp}} & \multirow{2}{*}{\begin{tabular}[c]{@{}c@{}}Gain over\\ G-PCC\end{tabular}} & \multicolumn{1}{c|}{\multirow{2}{*}{Bpp}} & \multirow{2}{*}{\begin{tabular}[c]{@{}c@{}}Gain over\\ G-PCC\end{tabular}} & \multicolumn{2}{c|}{$k$=16}                                                                                  & \multicolumn{2}{c|}{$k$=32}                                                                                           & \multicolumn{2}{c}{$k$=64}                                                                       \\ \cline{8-13} 
			\multicolumn{2}{c|}{}                                                         &                      & \multicolumn{1}{c|}{}                     &                                                                            & \multicolumn{1}{c|}{}                     &                                                                            & \multicolumn{1}{c|}{Bpp}  & \multicolumn{1}{c|}{\begin{tabular}[c]{@{}c@{}}Gain over\\ G-PCC\end{tabular}} & \multicolumn{1}{c|}{Bpp}           & \multicolumn{1}{c|}{\begin{tabular}[c]{@{}c@{}}Gain over\\ G-PCC\end{tabular}} & \multicolumn{1}{c|}{Bpp}           & \begin{tabular}[c]{@{}c@{}}Gain over\\ G-PCC\end{tabular} \\ \hline
			\multicolumn{1}{c|}{\multirow{5}{*}{10 bit}} & Thaidancer$\_$viewdep$\_$vox10 & 0.99                 & \multicolumn{1}{c|}{0.65}                 & -34.34\%                                                                   & \multicolumn{1}{c|}{0.65}                 & -34.34\%                                                                   & \multicolumn{1}{c|}{0.67} & \multicolumn{1}{c|}{-32.32\%}                                                  & \multicolumn{1}{c|}{\textbf{0.62}} & \multicolumn{1}{c|}{\textbf{-37.37\%}}                                         & \multicolumn{1}{c|}{\textbf{0.62}} & \textbf{-37.37\%}                                         \\
			\multicolumn{1}{c|}{}                        & boxer$\_$viewdep$\_$vox10      & 0.94                 & \multicolumn{1}{c|}{0.60}                 & -36.17\%                                                                   & \multicolumn{1}{c|}{0.59}                 & -37.23\%                                                                   & \multicolumn{1}{c|}{0.63} & \multicolumn{1}{c|}{-32.98\%}                                                  & \multicolumn{1}{c|}{\textbf{0.57}} & \multicolumn{1}{c|}{\textbf{-39.36\%}}                                         & \multicolumn{1}{c|}{0.58}          & -38.30\%                                                  \\
			\multicolumn{1}{c|}{}                        & loot$\_$vox10$\_$1200          & 0.97                 & \multicolumn{1}{c|}{0.63}                 & -35.05\%                                                                   & \multicolumn{1}{c|}{0.63}                 & -35.06\%                                                                   & \multicolumn{1}{c|}{0.66} & \multicolumn{1}{c|}{-31.96\%}                                                  & \multicolumn{1}{c|}{0.62}          & \multicolumn{1}{c|}{-36.08\%}                                                  & \multicolumn{1}{c|}{\textbf{0.61}} & \textbf{-37.11\%}                                         \\
			\multicolumn{1}{c|}{}                        & redandblack$\_$vox10$\_$1550   & 1.10                 & \multicolumn{1}{c|}{0.72}                 & -34.55\%                                                                   & \multicolumn{1}{c|}{0.74}                 & -32.73\%                                                                   & \multicolumn{1}{c|}{0.76} & \multicolumn{1}{c|}{-30.91\%}                                                  & \multicolumn{1}{c|}{0.71}          & \multicolumn{1}{c|}{-35.45\%}                                                  & \multicolumn{1}{c|}{\textbf{0.70}} & \textbf{-36.36\%}                                         \\
			\multicolumn{1}{c|}{}                        & Average                        & 1.00                 & \multicolumn{1}{c|}{0.65}                 & -35.00\%                                                                   & \multicolumn{1}{c|}{0.65}                 & -35.00\%                                                                   & \multicolumn{1}{c|}{0.68} & \multicolumn{1}{c|}{-32.00\%}                                                  & \multicolumn{1}{c|}{\textbf{0.63}} & \multicolumn{1}{c|}{\textbf{-37.00\%}}                                         & \multicolumn{1}{c|}{\textbf{0.63}} & \textbf{-37.00\%}                                         \\ \hline
			\multicolumn{1}{c|}{\multirow{3}{*}{9 bit}}  & Thaidancer$\_$viewdep$\_$vox9  & 0.99                 & \multicolumn{1}{c|}{0.64}                 & -35.35\%                                                                   & \multicolumn{1}{c|}{0.64}                 & -35.35\%                                                                   & \multicolumn{1}{c|}{0.67} & \multicolumn{1}{c|}{-32.32\%}                                                  & \multicolumn{1}{c|}{0.63}          & \multicolumn{1}{c|}{-36.36\%}                                                  & \multicolumn{1}{c|}{\textbf{0.60}} & \textbf{-39.39\%}                                         \\
			\multicolumn{1}{c|}{}                        & boxer$\_$viewdep$\_$vox9       & 0.96                 & \multicolumn{1}{c|}{0.60}                 & -37.50\%                                                                   & \multicolumn{1}{c|}{0.59}                 & -38.54\%                                                                   & \multicolumn{1}{c|}{0.62} & \multicolumn{1}{c|}{-35.42\%}                                                  & \multicolumn{1}{c|}{0.59}          & \multicolumn{1}{c|}{-38.54\%}                                                  & \multicolumn{1}{c|}{\textbf{0.56}} & \textbf{-41.67\%}                                         \\
			\multicolumn{1}{c|}{}                        & Average                        & 0.98                 & \multicolumn{1}{c|}{0.62}                 & -36.73\%                                                                   & \multicolumn{1}{c|}{0.62}                 & -36.73\%                                                                   & \multicolumn{1}{c|}{0.65} & \multicolumn{1}{c|}{-33.67\%}                                                  & \multicolumn{1}{c|}{0.61}          & \multicolumn{1}{c|}{-37.76\%}                                                  & \multicolumn{1}{c|}{\textbf{0.58}} & \textbf{-40.82\%}                                         \\ \hline
		\end{tabular}
	}
\end{table}

\subsubsection{Dense Object Point Clouds}
Besides, we also  test PCGs with higher geometry bit-depths and levels of sparsity which are referred to as the dense object point clouds. Since the SOTA {\it OctAttention}~\cite{fu2022octattention} does not include  this type of point clouds into the performance evaluation, we exclude it for comparative study below. We faithfully reproduce the result of {\it SparsePCGC}~\cite{wang2021sparse} and the anchor result of {\it G-PCC} is generated by the latest TMC13v14~\cite{tmc13code} as well.

\textbf{Dataset.} We select loot\_view\_vox12, redandblack\_viewdep\_vox12, longdress\_viewdep\_vox12, head\_00039\_vox12 and frog\_00067\_vox12  in {\it G-PCC} CTC~\cite{MPEG_GPCC_CTC} following the recommendation of ~\cite{MPEG_AI_PCC_SPLIT}  for training. The testing PCGs include soldier\_viewdep\_vox12, boxer\_viewdep\_vox12 and Thaidancer\_viewdep\_vox12. Note that PCGs mentioned above are all with 12-bit precision.

\textbf{Training and Testing.} In the training procedure, we train one model with scaling factor $S$ from 1 to $\frac{1}{8}$, and the number of neighbors $k$ is set as 16. In the testing procedure, we scale input PCGs with $S$ from 1 to $\frac{1}{32}$ to perform rate-distortion control when evaluating lossy compression performance. Other settings including evaluate metrics and loss function are the same as described in Sec. 4.2 as well.

\textbf{Results.} 
The proposed method not only shows great performance on the compression of LiDAR PGCs described in the main text, but also for these dense object PCGs. As shown in Table~\ref{tab:lossless_lossy_dense_PCG}, for lossless mode, our method achieves 50.18\% Bpp reduction against the {\it G-PCC} and 23.24\% Bpp saving over the {\it SparsePCGC}~\cite{wang2021sparse}. In lossy mode, our method offers more than 16\% and 17\% gains over the {\it G-PCC} in terms of D1 BD-rate and D2 BD-rate, which is around another 3\% BD-rate improvement when comparing with the {\it SparsePCGC}~\cite{wang2021sparse}.

\begin{table}[h]
	\centering
	\caption{Compression Performance Evaluation Using Dense Object PCGs for both Lossless and Lossy Scenarios. Anchor is standardized G-PCC using TMC13v14, and SparsePCGC is also provided.}
	\label{tab:lossless_lossy_dense_PCG}
	\resizebox{1\textwidth}{!}{
		\begin{tabular}{cc|cccc|cccc}
			\hline
			\multicolumn{2}{c|}{\multirow{3}{*}{Dense Object PCGs}}                              & \multicolumn{4}{c|}{Lossless (Bpp)}                                                                                                                                                                                             & \multicolumn{4}{c}{Lossy}                                                                                    \\ \cline{3-10} 
			\multicolumn{2}{c|}{}                                                         & \multicolumn{1}{c|}{\multirow{2}{*}{G-PCC}} & \multicolumn{1}{c|}{\multirow{2}{*}{SparsePCGC~\cite{wang2021sparse}}} & \multicolumn{1}{c|}{\multirow{2}{*}{Ours}} & \multirow{2}{*}{\begin{tabular}[c]{@{}c@{}}Gains over\\ G-PCC\end{tabular}} & \multicolumn{2}{c|}{D1-BD-rate}                                 & \multicolumn{2}{c}{D2-BD-rate}             \\ \cline{7-10} 
			\multicolumn{2}{c|}{}                                                         & \multicolumn{1}{c|}{}                       & \multicolumn{1}{c|}{}                            & \multicolumn{1}{c|}{}                      &                                                                             & \multicolumn{1}{c|}{SparsePCGC~\cite{wang2021sparse}} & \multicolumn{1}{c|}{Ours}     & \multicolumn{1}{c|}{SparsePCGC~\cite{wang2021sparse}} & Ours     \\ \hline
			\multicolumn{1}{c|}{\multirow{4}{*}{12 bit}} & soldier$\_$viewdep$\_$vox12    & \multicolumn{1}{c|}{3.86}                   & \multicolumn{1}{c|}{2.62}                        & \multicolumn{1}{c|}{2.00}                  & -48.19\%                                                                    & \multicolumn{1}{c|}{-16.07\%}   & \multicolumn{1}{c|}{-19.79\%} & \multicolumn{1}{c|}{-17.07\%}   & -20.07\% \\
			\multicolumn{1}{c|}{}                        & boxer$\_$viewdep$\_$vox12      & \multicolumn{1}{c|}{3.58}                   & \multicolumn{1}{c|}{2.15}                        & \multicolumn{1}{c|}{1.48}                  & -58.66\%                                                                    & \multicolumn{1}{c|}{-15.71\%}   & \multicolumn{1}{c|}{-19.15\%} & \multicolumn{1}{c|}{-16.60\%}   & -19.99\% \\
			\multicolumn{1}{c|}{}                        & Thaidancer$\_$viewdep$\_$vox12 & \multicolumn{1}{c|}{1.10}                   & \multicolumn{1}{c|}{0.78}                        & \multicolumn{1}{c|}{0.78}                  & -29.09\%                                                                    & \multicolumn{1}{c|}{-6.90\%}    & \multicolumn{1}{c|}{-9.76\%}  & \multicolumn{1}{c|}{-8.21\%}    & -11.21\% \\
			\multicolumn{1}{c|}{}                        & Average                        & \multicolumn{1}{c|}{2.85}                   & \multicolumn{1}{c|}{1.85}                        & \multicolumn{1}{c|}{1.42}                  & -50.18\%                                                                    & \multicolumn{1}{c|}{-12.89\%}   & \multicolumn{1}{c|}{-16.23\%} & \multicolumn{1}{c|}{-13.96\%}   & -17.09\% \\ \hline
		\end{tabular}
	}
\end{table}

\newpage
{
\small

\bibliographystyle{plain}
\bibliography{main}

}

\end{document}